\documentclass[10pt,twocolumn,letterpaper]{article}
\usepackage{pdfpages}
\usepackage{cvpr}              
\usepackage{graphicx}
\usepackage{amsmath}
\usepackage{amssymb}
\usepackage{booktabs}
\usepackage{bbm}
\usepackage{enumitem}
\usepackage{multirow}
\usepackage{tabularx}
\usepackage[normalem]{ulem}
\usepackage[table,xcdraw,dvipsnames]{xcolor}
\usepackage{bbding}
\usepackage{setspace}

\useunder{\uline}{\ul}{}
\usepackage[pagebackref,breaklinks,colorlinks]{hyperref}

\usepackage[capitalize]{cleveref}
\crefname{section}{Sec.}{Secs.}
\Crefname{section}{Section}{Sections}
\Crefname{table}{Table}{Tables}
\crefname{table}{Tab.}{Tabs.}

\def\cvprPaperID{5850} 
\def\confName{CVPR}
\def\confYear{2022}

\begin{document}

\title{Correlation Verification for Image Retrieval}
\author{Seongwon Lee \quad\quad Hongje Seong \quad\quad Suhyeon Lee \quad\quad Euntai Kim\thanks{Corresponding author.}\\
School of Electrical and Electronic Engineering, Yonsei University, Seoul, Korea\\
{\tt\small \{won4113, hjseong, hyeon93, etkim\}@yonsei.ac.kr}
}

\maketitle

\begin{abstract}
Geometric verification is considered a de facto solution for the re-ranking task in image retrieval. In this study, we propose a novel image retrieval re-ranking network named Correlation Verification Networks (CVNet). Our proposed network, comprising deeply stacked 4D convolutional layers, gradually compresses dense feature correlation into image similarity while learning diverse geometric matching patterns from various image pairs. To enable cross-scale matching, it builds feature pyramids and constructs cross-scale feature correlations within a single inference, replacing costly multi-scale inferences. In addition, we use curriculum learning with the hard negative mining and Hide-and-Seek strategy to handle hard samples without losing generality. Our proposed re-ranking network shows state-of-the-art performance on several retrieval benchmarks with a significant margin (+12.6\% in mAP on $\mathcal{R}$Oxford-Hard+1M set) over state-of-the-art methods. The source code and models are available online: \url{https://github.com/sungonce/CVNet}.

\end{abstract}
\vspace{-0.5cm}

\section{Introduction}
\label{sec:1.Introduction}
\vspace{-0.15cm}
Image retrieval is a long-standing problem in computer vision. This task aims to sort a database of images based on their similarities to the given query image.
For this task, global retrieval through global descriptor matching and geometric verification after local feature matching are mainly employed. These approaches typically comprise two primary components of the image retrieval framework that mutually complement one another. 
The global retrieval quickly performs a coarse retrieval across the database, and geometric verification re-ranks the coarse results by performing precise evaluation only on the potential candidates.
Along with deep learning, image retrieval has also advanced significantly.
In particular, several studies \cite{cao2020unifying, yang2021dolg, tan2021instance, noh2017large, teichmann2019detect,simeoni2019local} have been focused on extracting representative and distinctive features for global and local representations with deep learning.
However, geometric verification after local feature matching still plays an essential role in the re-ranking task in image retrieval, despite its drawbacks.
Owing to its \textit{verify-after-matching} structure, geometric verification is performed based on only sparse and thresholded feature correspondence. Moreover, it is neither learnable nor differentiable and requires iterative optimization even during testing.
In addition, geometric verification does not include a component that can handle multi-scale operation. Thus, several studies \cite{noh2017large,cao2020unifying,tan2021instance,philbin2007object} have attempted to solve the scale problem by repeating inference with the image pyramid to extract multi-scale local features. However, this is an extremely expensive process.

\begin{figure}[t]
\centering
\includegraphics[width=1.0\linewidth]{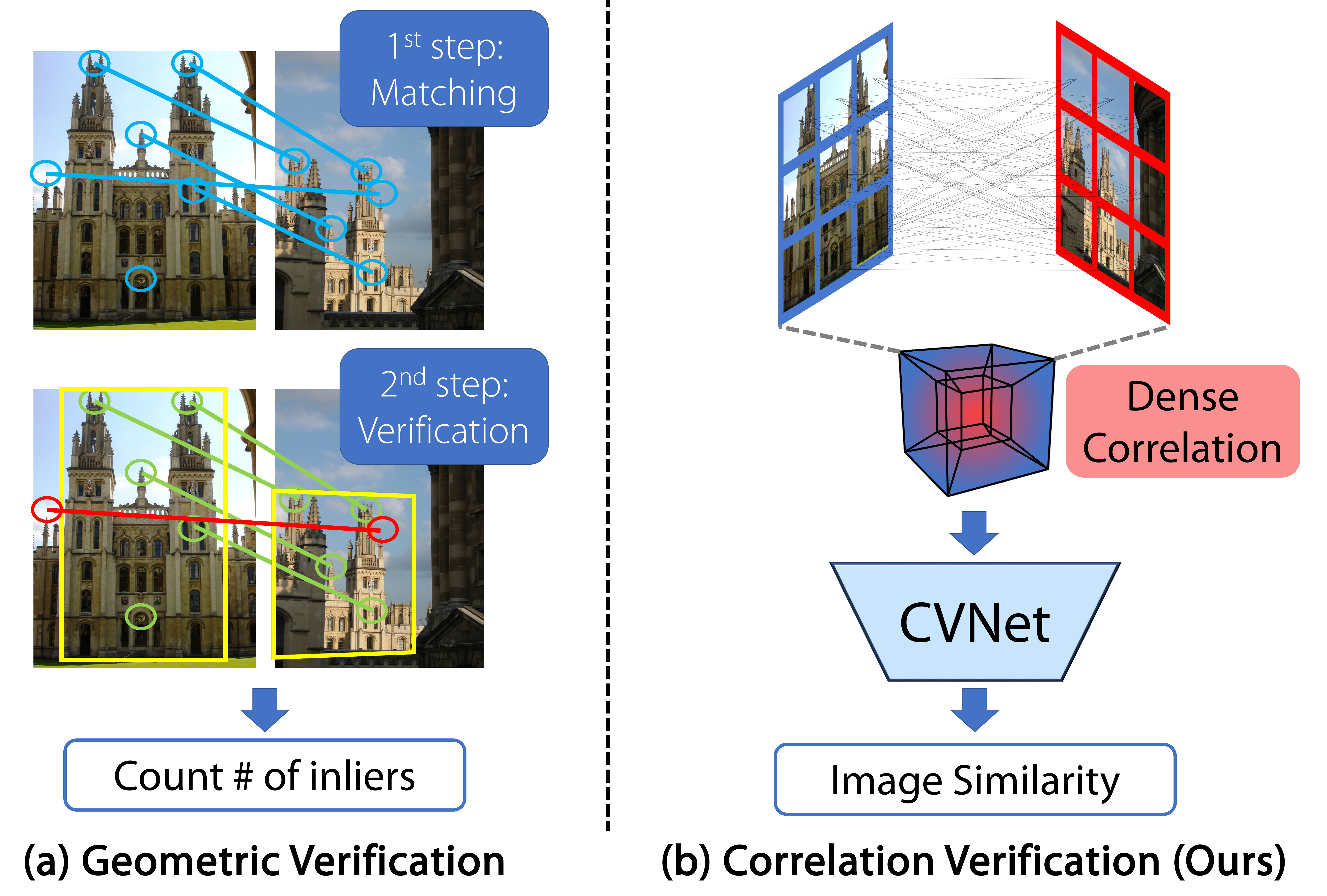}
\vspace{-0.6cm}
\caption{
Novel image retrieval re-ranking method named correlation verification that \textit{directly} predicts image similarity by leveraging dense feature correlation in a convolutional manner.
}
\vspace{-0.5cm}
\label{fig:fig1.introduce}
\end{figure}    

In this study, we propose an end-to-end learnable re-ranking network called \textit{Correlation Verification Networks} (CVNet) to replace the role of geometric verification in a better way.
The proposed network \textit{directly} evaluates semantic and geometric relations by leveraging dense feature correlations in a convolutional manner.
Following the successful architectural design of representative 2D convolutional neural networks (CNN), we design a 4D CNN with a pyramid structure of deeply stacked 4D convolution layers.
It compresses the correlation between semantic cues into image similarity while learning diverse geometric matching patterns from a large number of image pairs.
To ensure robustness even for large scale difference problems, it expands the single-scale feature to a feature pyramid for each image, forming cross-scale correlations between feature pyramids.
This structure enables cross-scale matching with a single inference while replacing the multi-scale inference conventionally used in image retrieval.
Our model does not require additional inference to extract local information; therefore the feature extraction latency, which significantly affects online retrieval time, is considerably reduced compared with other re-ranking methods.
Similar to several computer vision problems, image retrieval suffers from the problem of hard samples. We address these challenges through curriculum learning using the hard negative mining and Hide-and-Seek \cite{singh2017hide} strategy in the training phase. This improves the overall performance by focusing on hard samples without losing generality in the case of normal ones.
Our proposed re-ranking network shows state-of-the-art performance on several image retrieval benchmarks with a significant margin over several state-of-the-art methods. Our main contributions are as follows:
\vspace{-0.15cm}
\begin{itemize}[leftmargin=5mm]
  \setlength{\itemsep}{1pt}
  \setlength{\parskip}{0pt}
  \setlength{\parsep}{0pt}
    \item We present Correlation Verification Networks (CVNet), which is a powerful re-ranking model that directly predicts the similarity of an image pair based on dense feature correlation.
    \item To replace expensive multi-scale inference, we construct cross-scale correlations within the model and perform cross-scale matching using a  single inference.
    \item We propose curriculum learning using the hard negative mining and Hide-and-Seek strategy to handle hard samples without losing generality.
    \item The proposed model achieves new state-of-the-art performance on several image retrieval benchmarks: $\mathcal{R}$Oxford (+1M), $\mathcal{R}$Paris (+1M), and GLDv2-retrieval. 
\end{itemize}

\vspace{-0.15cm}
\section{Related Work}
\label{sec:2.Related_Work}
\vspace{-0.15cm}

\paragraph{Image retrieval.}

Over the past few decades, image retrieval\cite{sivic2003video,jegou2010aggregating,jegou2011aggregating,radenovic2016cnn,arandjelovic2016netvlad, teichmann2019detect,radenovic2018fine,cao2020unifying} has been one a primary focus of computer-vision studies.
In pioneering research, handcrafted local features \cite{lowe2004distinctive, bay2008speeded} have been employed for global retrieval and re-ranking. A global retrieval with a global descriptor that aggregates handcrafted local features \cite{sivic2003video, philbin2007object, philbin2008lost, jegou2008hamming, jegou2010aggregating, jegou2011aggregating} is performed first, and spatial verification \cite{philbin2007object, philbin2008lost, avrithis2014hough} via local feature matching with RANSAC \cite{fischler1981random} is performed to re-rank putative retrieval results. Afterward, with the advancements in deep learning, global \cite{babenko2014neural, babenko2015aggregating, arandjelovic2016netvlad, gordo2017end, radenovic2018fine, tolias2015particular, cao2020unifying, yang2021dolg} and local features \cite{barroso2019key, dusmanu2019d2, luo2019contextdesc, mishchuk2017working, mishkin2018repeatability,noh2017large,yi2016lift, cao2020unifying} extracted from deep-learning networks have replaced handcrafted features.

Although the techniques of global and local representations has progressed significantly, geometric verification remains a de facto solution for image retrieval re-ranking in both conventional \cite{philbin2007object,philbin2008lost, xu2012learning} and recent studies \cite{noh2017large,cao2020unifying,simeoni2019local,teichmann2019detect}. In a recent study, Reranking Transformers (RRT) \cite{tan2021instance} were proposed as a replacement for geometric verification by leveraging the transformer structure \cite{vaswani2017attention}. However, no significant improvement in performance was reported. In this study, we propose a novel re-ranking solution that exhibits powerful retrieval performance.

\vspace{-0.4cm}
\paragraph{Diffusion / Query expansion.}
Among the re-ranking methods, several methods such as diffusion \cite{iscen2017efficient, chang2019explore} and query expansion \cite{chum2007total, radenovic2018fine} exist that require additional expenses to traverse the entire database. However, because this study focuses on improving image matching for single pairs, we do not consider these re-ranking methods.

\vspace{-0.4cm}
\paragraph{4D convolutional neural network.}
4D convolution is a promising solution that has received considerable attention for tasks that require interpretation of the relationship between two images (\eg visual dense correspondence prediction \cite{rocco2018neighbourhood, min2021convolutional, yang2019volumetric, li2020correspondence} and few-shot segmentation \cite{min2021hypercorrelation}). 
The primary difference between the aforementioned tasks and image retrieval is that the former aims for a 2D (single image side) \cite{min2021hypercorrelation} or 4D (both image sides) \cite{min2021convolutional,min2021convolutional,yang2019volumetric} dense output, whereas the latter requires a single similarity value.
Therefore, in this study, we propose a novel structure that gradually compresses the 4D feature correlation through deeply stacked 4D convolution layers.

\begin{figure*}[t]
\centering
\includegraphics[width=1.0\linewidth]{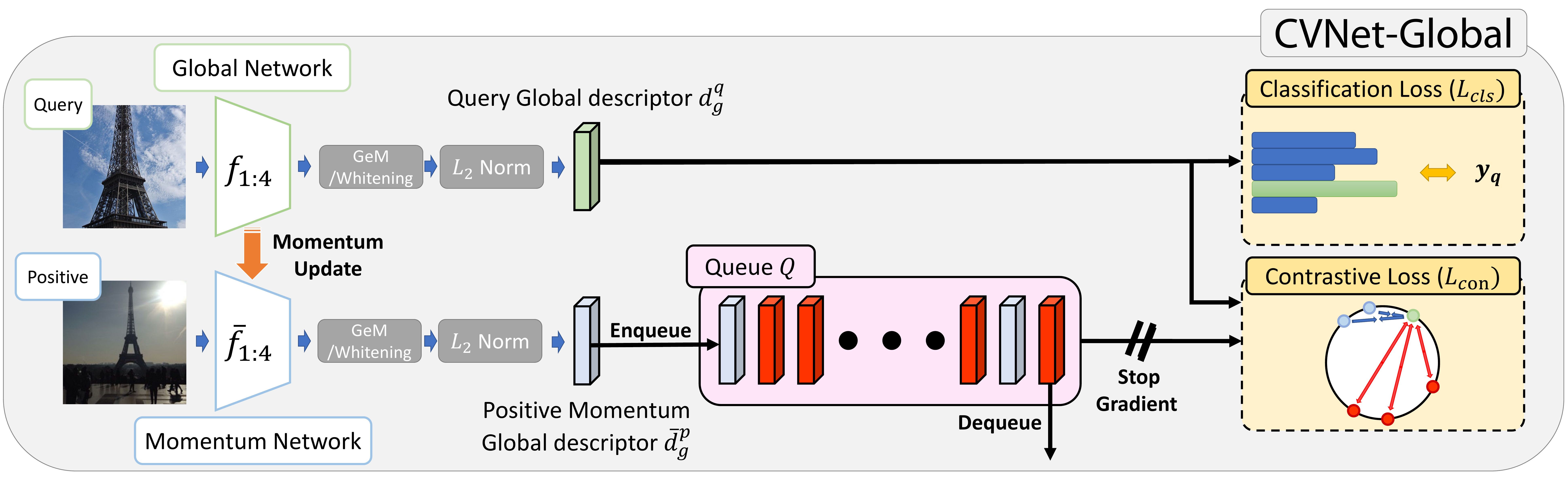}
\vspace{-0.6cm}
\caption{
Illustration of the proposed Global backbone network (CVNet-Global) and its training objective. The network has two objectives: classification loss and contrastive loss. To utilize several samples without a computational burden in contrastive learning, momentum network and queue structure are adopted from MoCo \cite{he2020momentum}. The combination of these objectives enables the network to learn intra-class variability and inter-class distinctiveness, which is required for image retrieval task.
}
\vspace{-0.5cm}
\label{fig:CVNet_Global}
\end{figure*}

\vspace{-0.4cm}
\paragraph{Hide-and-Seek.}
Hide-and-Seek \cite{singh2017hide} is an augmentation technique that has been proposed to improve object localization performance in weakly supervised fields. To address the drawback that the network focuses only on the most salient areas, a few random patches of the image are masked to induce the network to make robust predictions despite having visual access only to less salient areas. We found that the Hide-and-Seek approach could improve the image retrieval performance by enabling accurate matching even on hard samples, such as those involving occlusion or truncation. 
In this study, we apply Hide-and-Seek to our model in a curriculum manner to ensure robustness when handling hard samples without losing generality.

\vspace{-0.15cm}
\section{Global Backbone Network (CVNet-Global)}
\label{sec:3.Global_Backbone_Network}
\vspace{-0.15cm}

In this section, we introduce our proposed global backbone network named CVNet-Global. An overview of CVNet-Global is shown in \cref{fig:CVNet_Global}.
Our proposed global backbone network, that takes a single image $\mathbf{I} \in \mathbb{R}^{3 \times H \times W}$ as the input, is used to extract the global descriptor $\mathbf{d}_g \in \mathbb{R}^{C_g}$ for global image retrieval and local feature map $\mathbf{F} \in \mathbb{R}^{C_l \times H_l \times W_l}$ for the re-ranking phase.
We adopt multi-objective loss\cite{berman2019multigrain} that jointly optimizes the classification loss and contrastive loss to induce the network to learn more distinctive and robust global and local representations.

\vspace{-0.15cm}
\subsection{Structure}
\label{sec:3.1.Structure}
\vspace{-0.15cm}

Inspired by the momentum-contrastive structure of MoCo \cite{he2020momentum}, we build two networks: the global backbone network $f$ and its momentum network $\bar{f}$. These two networks are based on ResNet \cite{he2016deep}. $f_i$ denotes $i$th \textit{ResBlock}. Global Average Pooling is replaced with learnable GeM pooling \cite{radenovic2016cnn} with power initialized to 3.0, and a whitening FC layer\cite{gordoa2012leveraging} and L2-normalization are added after the pooling layer. We build a queue $\mathbf{Q} \in \{\bar{\mathbf{d}}_g^i\}_{i=1}^K$, to save momentum global descriptors for each iteration and utilize them as contrastive samples. 

\vspace{-0.10cm}
\subsection{Training Objective}
\label{sec:3.2.Training_Objective}
\vspace{-0.10cm}

\paragraph{Classification loss.}
At each iteration, the query image $\mathbf{I}_q$ is fed into the global network $f$ to compute the query global descriptor $\mathbf{d}^q_g$. With $\mathbf{d}^q_g$, CurricularFace \cite{huang2020curricularface}-margined classification loss $\mathcal{L}_{cls}$ is computed as
\vspace{-0.15cm}
\begin{equation}
{
\mathcal{L}_{cls} =  - \log\frac{\exp ( \mathcal{C} ({\mathbf{W}_{y_g}}^T \mathbf{d}^q_g, 1)/\tau )}{{\sum}_{i=1}^{N}{\exp (\mathcal{C} ({{\mathbf{W}_{y_i}}^T \mathbf{d}^q_g, \mathbbm{1}^i_q})/ \tau )}},
}
\vspace{-0.15cm}
\label{eq:Global_Classification_Loss}
\end{equation}
where $\mathbf{W}$ is the class weight, $\tau$ is the scale parameter, $y_g$ is the ground-truth class, and $\mathbbm{1}^i_q$ is an indicator that shows whether the $i$th class $y_i$ is identical to $y_g$. $\mathcal{C}$ is a function that adds a CurricularFace margin to cosine similarity with its margin term $m$.

\begin{figure*}[t]
\centering
\includegraphics[width=1.0\linewidth]{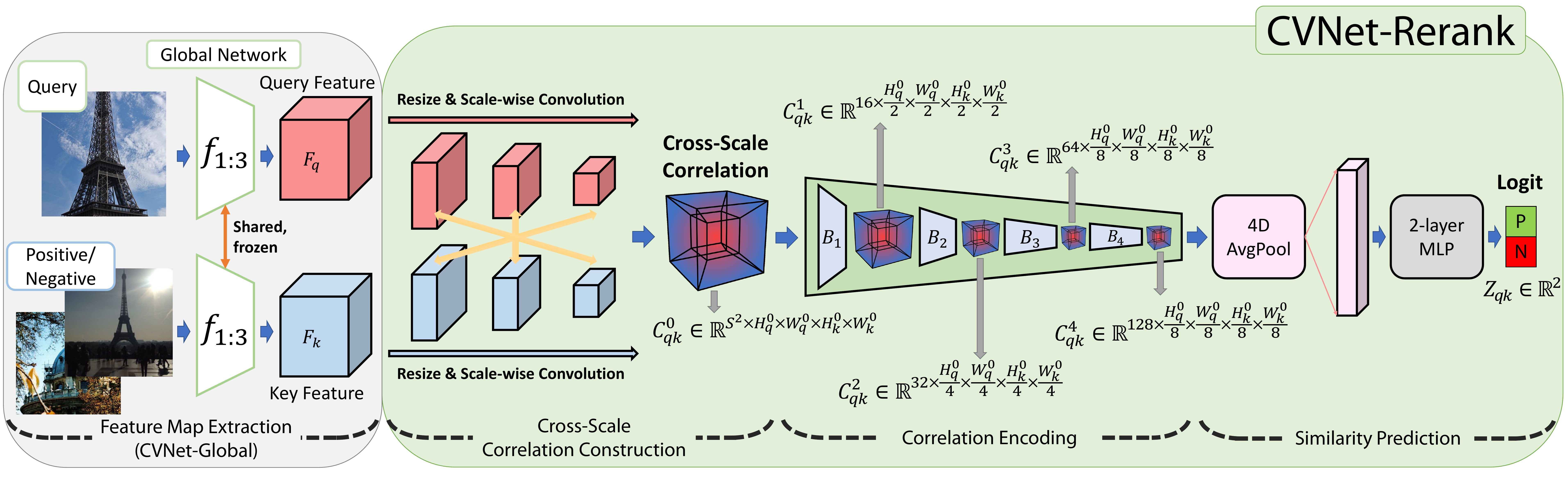}
\vspace{-0.6cm}
\caption{
Illustration of the proposed Re-ranking network (CVNet-Rerank). The proposed network takes pair of feature maps extracted from the trained CVNet-Global model as input, constructs a cross-scale feature correlation, and gradually compresses it to image similarity of a pair with deeply stacked 4D convolution layers.
}
\label{fig:CVNet_Rerank}
\vspace{-0.5cm}
\end{figure*}

\vspace{-0.40cm}
\paragraph{Momentum contrastive loss.}
At each iteration, a positive image $\mathbf{I}_p$ with the same label as the query image $\mathbf{I}_q$ is sampled and fed into the momentum network $\bar{f}$ to compute the positive momentum global descriptor $\bar{\mathbf{d}}^p_g$. The descriptor $\bar{\mathbf{d}}^p_g$ is updated to queue $\mathbf{Q}$ while dequeuing the last element of the queue. Then, queue $\mathbf{Q}$ holds at least one momentum sample with the same label as the query including $\bar{\mathbf{d}}^p_g$. Thus we use the CurriculurFace-margined momentum contrastive loss $\mathcal{L}_{con}$:
\vspace{-0.15cm}
\begin{equation}
{
\mathcal{L}_{con}\! = \!\frac{-1}{\left | P(q) \right |} \sum\limits_{p \in P(q)}\!\!\log\!\frac{ \exp \left( \bar{\mathcal{C}} \left( \mathbf{d}_g^q \cdot \bar{\mathbf{d}}_g^p, 1 \right)\! / \tau \right)}{\sum\limits_{i \in \{\!p\!\}\!\bigcup\!N\!(\!q\!)}\exp \left( \bar{\mathcal{C}} \left( \mathbf{d}_g^q \cdot \bar{\mathbf{d}}_g^i, \mathbbm{1}^i_q \right)\! / \tau \right)},
}
\label{eq:Global_Contrastive_Loss}
\end{equation}
where $\bar{\mathcal{C}}$ is identical to $\mathcal{C}$, but updates its moving average parameter separately with $\mathcal{C}$. $P(q)$ and $N(q)$ are the in-queue positive and negative set, respectively.

\vspace{-0.45cm}
\paragraph{Total loss.}
Finally, the total loss of our global backbone network $\mathcal{L}_g$ is the weighted sum of the classification loss $\mathcal{L}_{cls}$ and contrastive loss $\mathcal{L}_{con}$:
\vspace{-0.15cm}
\begin{equation}
{
\mathcal{L}_g = \lambda_{cls}\mathcal{L}_{cls} + \lambda_{con}\mathcal{L}_{con}.
}
\label{eq:Global_Total_Loss}
\vspace{-0.15cm}
\end{equation}

Note that, optimizer only updates the global backbone network $f$. The momentum network $\bar{f}$ is momentum updated with a momentum of $\eta$.

\vspace{-0.10cm}
\section{Re-Ranking Network (CVNet-Rerank)}
\label{sec:4.Re_Ranking_Network}
\vspace{-0.10cm}
In this section, we introduce our proposed re-ranking network, named CVNet-Rerank.
An overview of CVNet-Rerank is shown in \cref{fig:CVNet_Rerank}.
Our proposed re-ranking network, which takes a pair of local feature maps $(\mathbf{F}_q, \mathbf{F}_k)$ of images $(\mathbf{I}_q, \mathbf{I}_k)$ as input, is used to predict the similarity $s_l^{q,k} \in \mathbb{R}^1$ between two images. It subsequently re-ranks the global image retrieval results based on the results of the predicted similarity. The local feature maps $(\mathbf{F}_q, \mathbf{F}_k)$ are extracted from the intermediate layer of the global backbone network $f$, that is fully trained and frozen.
Representative 2D CNN architectures (\eg VGG \cite{simonyan2015very} and ResNet \cite{he2016deep}) stack several 2D convolutional layers, followed by spatial-dimensional down-sampling to capture diverse level features in an image and compress it to fine-grained information. Inspired by the aforementioned structure, the proposed re-ranking network gradually compresses the feature correlation with deeply stacked 4D convolution layers and predicts the image similarity using the classifier. 

\vspace{-0.10cm}
\subsection{Cross-scale Correlation Construction}
\label{sec:4.1.Cross_scale_Correlation_Construction}
\vspace{-0.10cm}
Because image retrieval must be robust for scale difference, several image retrieval methods that use local features built a multi-scale local feature set through multiple inferences using an image pyramid.
Here, following \cite{min2021convolutional}, we expand the extracted feature map to a multi-scale feature pyramid to capture semantic cues from different scales inside the model, thus avoiding the expensive task of multi-scale inference.
Given a pair of query and key images $\mathbf{I}_q, \mathbf{I}_k \in \mathbb{R}^{3 \times H \times W}$, we extract the local feature maps $\mathbf{F}_q, \mathbf{F}_k \in \mathbb{R}^{C_l \times H_l \times W_l}$ using the global backbone network $f$. After feature extraction, we construct a feature pyramid $\{\mathbf{F}^s\}^S_{s=1}$,where $S$ is the number of scales, by repeatedly resizing the extracted feature map $F$ with a scaling factor of $1/\sqrt{2}$. Each level of the feature pyramid passes the scale-wise $3\times3$ convolution layer, thereby reducing the channel dimension of each layer to $C'_l$ to capture semantic information with diverse receptive field sizes while reducing the memory footprint of our image retrieval framework.
With the constructed query feature pyramid $\{\mathbf{F}_q^s\}^S_{s=1}$ and key feature pyramid $\{\mathbf{F}_k^s\}^S_{s=1}$, we compute a 4-dimensional cross-scale correlation set $\{\mathbf{C}_{qk}^{s_q,s_k}\}^{(S,S)}_{(s_q,s_k)=(1,1)}$ of size $S^2$ using cosine similarity and ReLU function:
\begin{equation}
\mathbf{C}_{qk}^{s_q,s_k}(\mathbf{p}_q, \mathbf{p}_k) = \text{ReLU} \left (\frac{\mathbf{F}_q^{s_q}(\mathbf{p}_q) \cdot  \mathbf{F}_k^{s_k}(\mathbf{p}_k)}{\left \| \mathbf{F}_q^{s_q}(\mathbf{p}_q) \right \| \left \| \mathbf{F}_k^{s_k}(\mathbf{p}_k)  \right \|}  \right ),
\label{eq:Correlation_Computation}
\end{equation}
where $\mathbf{p}_q$ and $\mathbf{p}_k$ are the pixel positions in each feature map. Finally, we interpolate all the correlations to obtain the original feature resolution $H_l \times W_l$ for each image side, stack all the correlations, and construct a cross-scale correlation set $\mathbf{C}^0_{qk} \in \mathbb{R}^{S^2 \times H_l \times W_l \times H_l \times W_l}$.

\begin{figure}[t]
\centering
\includegraphics[width=1.0\linewidth]{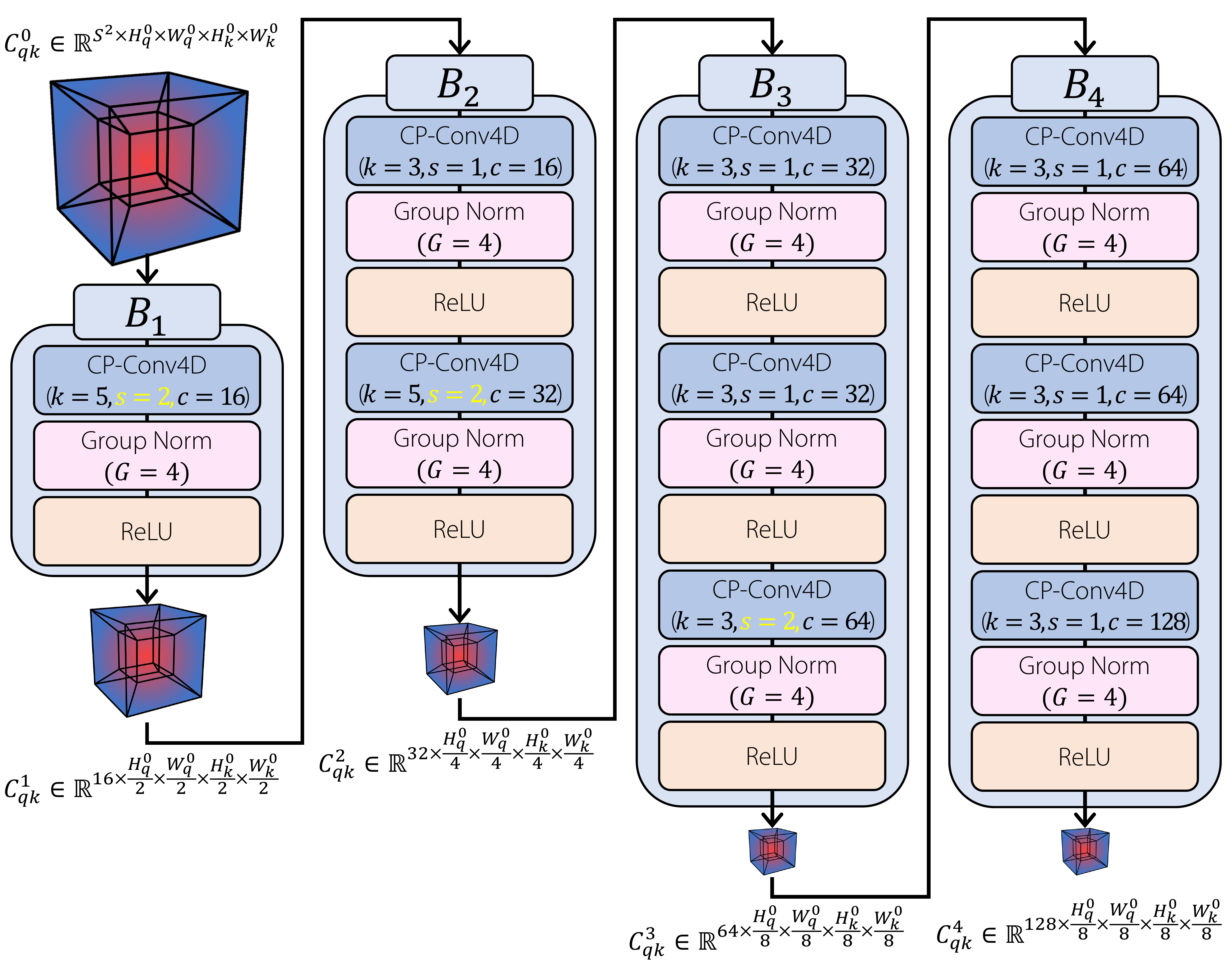}
\vspace{-0.6cm}
\caption{
The detailed structure of the proposed 4D correlation Encoder. The proposed encoder structure gradually compresses the cross-scale correlation into a fine-grained correlation cue.
}
\label{fig:fig4.4D_encoder_structure}
\vspace{-0.5cm}
\end{figure}

\subsection{4D Correlation Encoder}
\label{sec:4.2.4D_Correlation_Encoder}
\vspace{-0.15cm}
Our correlation encoder takes the cross-scale correlation set $\mathbf{C}^0_{qk} \in \mathbb{R}^{S^2 \times H_l \times W_l \times H_l \times W_l}$ and gradually compresses it into a binary class logit $\mathbf{Z}_{qk}=\{z_0, z_1\} \in \mathbb{R}^2$.
We construct our encoder with a sequence of 4D convolution blocks, followed by a global average pooling layer and a 2-layer MLP classifier. Except for the last 4D convolution block, the remaining blocks perform spatial dimension down-sampling by constructing each last convolutional layer as a stride convolution. Na\"ive 4D convolution is computationally intensive and, therefore, unsuitable for online re-ranking. Using the knowledge taken from findings of previous studies, we adopt a center-pivot 4D convolution \cite{min2021hypercorrelation} to reduce the burden of using high-dimensional kernels and enable real-time image re-ranking.
With this pyramid structure of 4D convolution, the cross-scale feature correlation set is encoded as a fine-grained correlation cue $\mathbf{C}^{1:4}_{qk}$. It is subsequently converted into a class logit $\mathbf{Z}_{qk}$ through spatial dimension average pooling and a binary classifier.

\subsection{Training Objective}
\label{sec:4.3.Training_Objective}
\vspace{-0.15cm}

Our re-ranking network is trained to minimize the cross-entropy loss for query and key pair $\left(q,k\right)$:
\vspace{-0.15cm}
\begin{equation}
\mathcal{L}_{r}^{qk} = \mathbf{CE}(\mathbf{Softmax}(\mathbf{Z}_{qk}), \mathbbm{1}^k_q). 
\end{equation}
\label{eq:Rerank_Loss}
\vspace{-0.5cm}

We symmetrically convert the loss $\mathcal{L}_r^{qk}$ to $\mathcal{L}_r^{kq}$ by reversing the query-key position. Afterward, we apply them to positive $p$ and negative key samples $n$, respectively. The final loss for our re-ranking network is constructed as follows:
\vspace{-0.15cm}
\begin{equation}
\mathcal{L}_{r} = \left(\mathcal{L}_{r}^{qp} + \mathcal{L}_{r}^{pq} + \mathcal{L}_{r}^{qn} + \mathcal{L}_{r}^{nq}\right)/4 . 
\end{equation}
\label{eq:Rerank_Total_Loss}
\vspace{-0.6cm}

\vspace{-0.15cm}
\subsection{Training with Hard Samples}
\label{sec:4.4:Training_with_Hard_Samples}
\vspace{-0.15cm}
Because image re-ranking is performed on images that look similar at first glance, it must be robust against hard samples. Thus, we propose a method to train a network by focusing on hard samples through hard negative mining and Hide-and-Seek augmentation. Although hard samples are beneficial for model training, a possibility of losing generality in the case of normal samples exists. Carefully considering this concern, we apply hard negative mining and Hide-and-Seek augmentation in a curriculum learning manner to train the re-ranking network to make more accurate predictions without losing generality in the case of normal ones while concentrating on hard samples. 

\begin{figure}[t]
\centering
\includegraphics[width=1.0\linewidth]{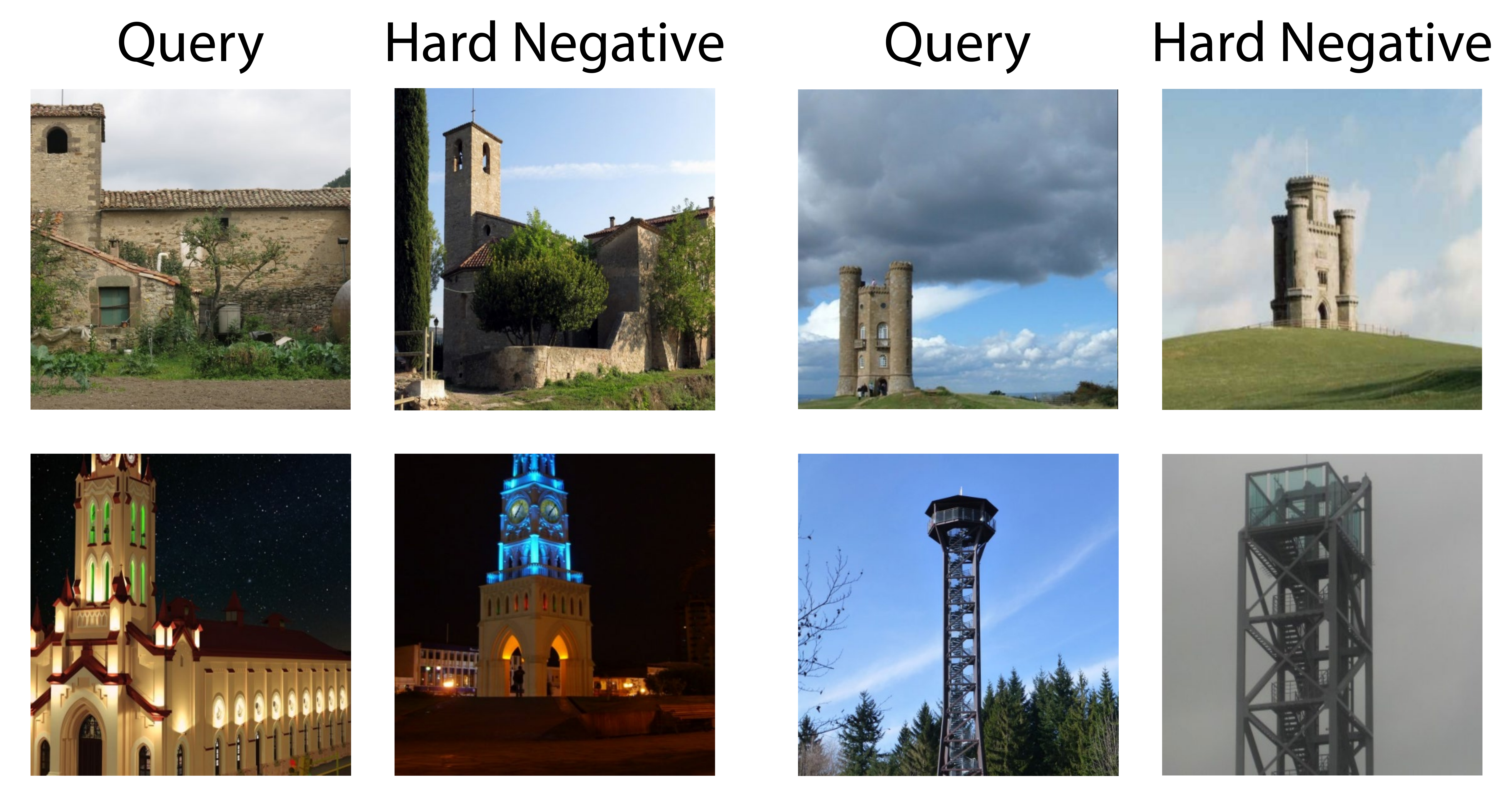}
\vspace{-0.6cm}
\caption{
Examples of the query and hard negative samples of the GLDv2-clean dataset. These pairs look similar at first glance, but a closer look reveals several differences.
}
\label{fig:fig5.hard_negative_mining}
\vspace{-0.5cm}
\end{figure}

\vspace{-0.45cm}
\paragraph{Hard negative mining.}
We selected hard-negative samples with help of trained global descriptors. For every sample in the training dataset, the top 10 negatives are selected in order of the highest global descriptor matching score. Example results of hard negative mining are shown in \cref{fig:fig5.hard_negative_mining}.

\begin{figure}[t]
\centering
\includegraphics[width=1.0\linewidth]{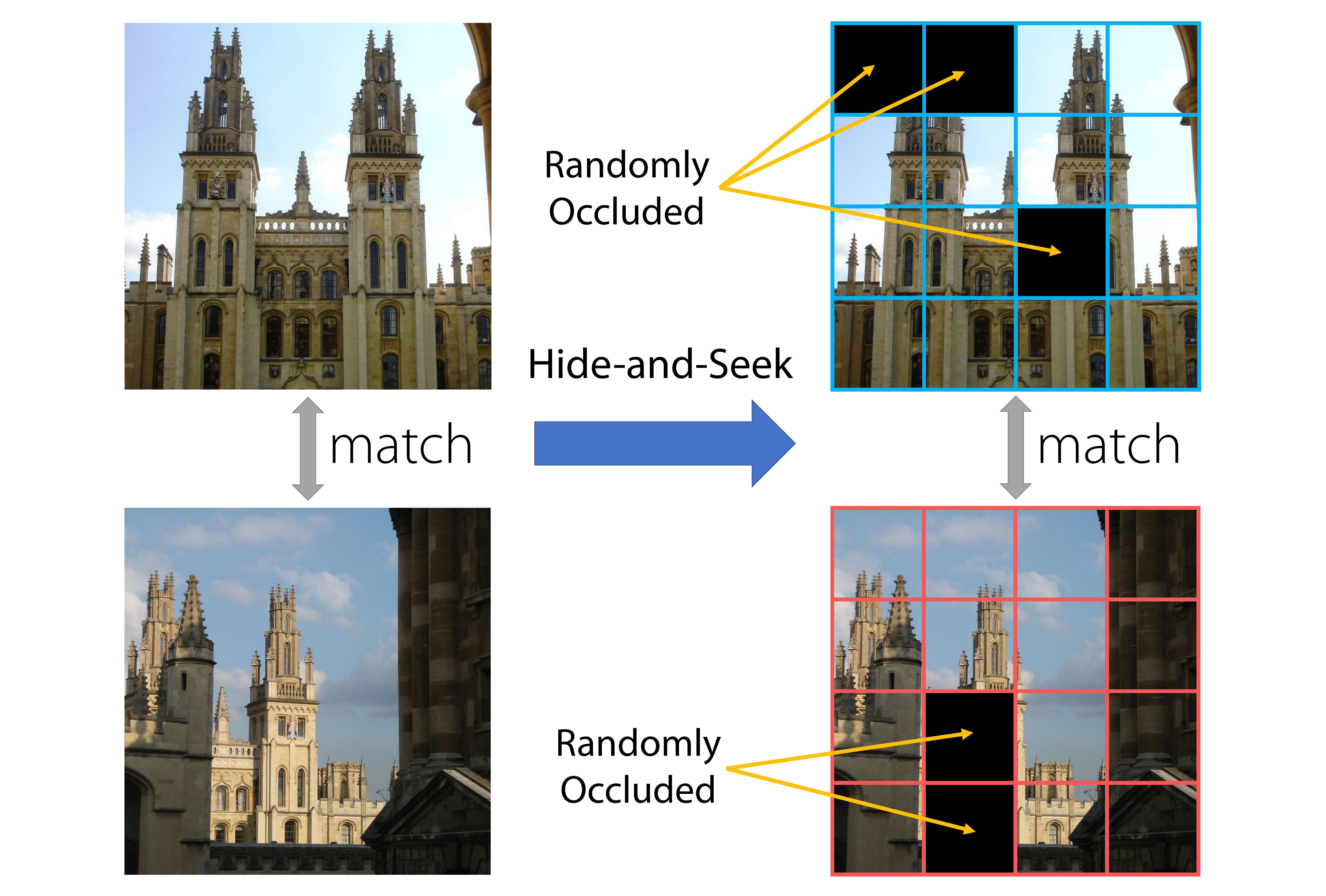}
\vspace{-0.6cm}
\caption{
With Hide-and-Seek, the re-ranking network can effectively learn hard-matching cases by randomly hiding parts of matching pairs to give images an occlusion-like effect.
}
\label{fig:fig6.hide_and_seek}
\vspace{-0.5cm}
\end{figure}

\vspace{-0.45cm}
\paragraph{Hide-and-Seek.}
Similar to several computer vision studies, occlusion is a primary obstacle in image retrieval tasks. To solve this problem, we apply Hide-and-Seek \cite{singh2017hide}  augmentation to synthetically generate matching situations that involve occlusions. In the original Hide-and-Seek method, the input image is divided into grids, and probabilistic deactivation is applied to each grid section.
Similarly, we randomly deactivate each pixel value from each input feature map. This can have an effect similar to that of applying occlusion to the receptive field of the original image that corresponds to one pixel in the feature map.
This concept is illustrated in \cref{fig:fig6.hide_and_seek}.

\begin{table*}[t]
\small
\centering
\setlength{\extrarowheight}{-4.5pt}
\addtolength{\extrarowheight}{\aboverulesep}
\addtolength{\extrarowheight}{\belowrulesep}
\setlength{\aboverulesep}{0pt}
\setlength{\belowrulesep}{0pt}
\resizebox{0.9\linewidth}{!}{\begin{tabular}{lcccclcccclcc}
\toprule
\multicolumn{1}{c}{\multirow{2}{*}{Method}} & \multicolumn{4}{c}{Medium} &           & \multicolumn{4}{c}{Hard}  &  & \multicolumn{2}{c}{Multi-scale} \\ \cmidrule(l){2-5} \cmidrule(l){7-10} \cmidrule(l){12-13}  
\multicolumn{1}{c}{}                        & $\mathcal{R}$Oxf  & +1M  & $\mathcal{R}$Par & +1M  &           & $\mathcal{R}$Oxf & +1M  & $\mathcal{R}$Par & +1M  &  & global          & local         \\ \midrule
\multicolumn{13}{l}{\textit{\textbf{(A) Local feature aggregation (+ Local feature re-ranking)}}}                                                                                                    \\
DELF-ASMK*+SP \cite{noh2017large, radenovic2018revisiting}                               & 67.8  & 53.8 & 76.9 & 57.3 &           & 43.1 & 31.2 & 55.4 & 26.4 &  - & 7          \\
DELF-D2R-R-ASMK* (GLDv1) \cite{teichmann2019detect}                       & 73.3  & 61.0 & 80.7 & 60.2 &           & 47.6 & 33.6 & 61.3 & 29.9 &  & - & 7           \\
\,\;+ SP (Rerank Top-100) \cite{teichmann2019detect}                       & 76.0  & 64.0 & 80.2 & 59.7 &           & 52.4 & 38.1 & 58.6 & 29.4 &  & - & 7           \\
R50-How-ASMK,n=2000 \cite{tolias2020learning}                        & 79.4  & 65.8 & 81.6 & 61.8 &           & 56.9 & 38.9 & 62.4 & 33.7 &  & - & 7           \\ \midrule
\multicolumn{13}{l}{\textit{\textbf{(B) Global features (+ Local feature re-ranking)}}}                                                                                 \\
R101-GeM$^\uparrow$ \cite{radenovic2018fine,simeoni2019local}                         & 65.3  & 46.1 & 77.3 & 52.6 &           & 39.6 & 22.2 & 56.6 & 24.8 &  & 3               & -             \\
\,\;+DSM (Rerank Top-100) \cite{simeoni2019local}                        & 65.3  & 47.6 & 77.4 & 52.8 &           & 39.2 & 23.2 & 56.2 & 25.0 &  & 3               &  3             \\
R101-GeM-AP (GLDv1) \cite{revaud2019learning}                         & 66.3  & -    & 80.2 & -    &           & 42.5 & -    & 60.8 & -    &  & 1               & -             \\
R101-GeM+SOLAR (GLDv1) \cite{ng2020solar} & 69.9  & 53.5 & 81.6 & 59.2 &           & 47.9 & 29.9 & 65.5 & 33.4 &  & 3               & -             \\
R50-DELG (Global-only, GLDv2-clean) \cite{cao2020unifying}                      & 73.6  & 60.6 & 85.7 & 68.6 &           & 51.0 & 32.7 & 71.5 & 44.4 &  & 3               & -             \\
\,\;+ GV (Rerank Top-100) \cite{cao2020unifying}                       & 78.3  & 67.2 & 85.7 & 69.6 &           & 57.9 & 43.6 & 71.0 & 45.7 &  & 3               & 7             \\
\,\;+ GV (Rerank Top-200) \cite{cao2020unifying,tan2021instance}       & 79.2  & 68.2 & 85.5 & 69.6 &           & 57.5 & 42.9 & 67.2 & 44.5 &  & 3               & 7             \\
\,\;+ RRT (Rerank Top-100) \cite{tan2021instance}                      & 78.1  & 67.0 & 86.7 & 69.8 &           & 60.2 & 44.1 & 75.1 & 49.4 &  & 3               & 7             \\
\,\;+ RRT (Rerank Top-200) \cite{tan2021instance}                      & 79.5  & 68.6 & 87.8 & 71.5 &           & \uline{62.5} & 46.3 & 77.1 & 52.3 &  & 3               & 7             \\
R101-DELG (Global-only, GLDv2-clean) \cite{cao2020unifying}                     & 76.3  & 63.7 & 86.6 & 70.6 &           & 55.6 & 37.5 & 72.4 & 46.9 &  & 3               & -             \\
\,\;+ GV (Rerank Top-100) \cite{cao2020unifying}                       & 81.2  & 69.1 & 87.2 & 71.5 &           & 64.0 & 47.5 & 72.8 & 48.7 &  & 3               & 7             \\
\,\;+ RRT (Rerank Top-100) \cite{cao2020unifying}                       & 79.9  & - & 87.6 & - &           & \uline{64.1} & - & 76.1 & - &  & 3               & 7             \\
\,\;+ SuperGlue (Rerank Top-100) \cite{cao2020unifying,sarlin2020superglue}                       & 79.7  & - & 87.1 & - &           & 62.1 & - & 71.5 & - &  & 3               & 7             \\

R50-DOLG (GLDv2-clean) \cite{yang2021dolg}                      & \uline{80.5}  & \uline{76.6} & \uline{89.8} & \uline{80.8} &           & 58.8 & \uline{52.2} & \uline{77.7} & \uline{62.8} &  & \multicolumn{2}{c}{5}           \\
R101-DOLG (GLDv2-clean) \cite{yang2021dolg}                     & \uline{81.5}  & \uline{77.4} & \uline{91.0} & \uline{83.3} &           & 61.1 & \uline{54.8} & \uline{80.3} & \uline{66.7} &  & \multicolumn{2}{c}{5}           \\ \midrule
\multicolumn{13}{l}{\textit{\textbf{(C) Ours}}}                                                                                                                         \\
\textbf{R50-CVNet-Global (GLDv2-clean)}     & 81.0  & 72.6 & 88.8 & 79.0 &           & 62.1 & 50.2 & 76.5 & 60.2 &  & 3               & -             \\
\,\;\textbf{+ CVNet-Rerank (Rerank Top-100)}              & 86.1  & 77.6 & 89.4 & 79.9 &           & 72.8 & 61.1 & 78.6 & 63.9 &  & 3               & 1             \\
\,\;\textbf{+ CVNet-Rerank (Rerank Top-200)}              & 87.2  & 78.9 & 90.0 & 81.2 &           & 74.5 & 62.9 & 79.5 & 66.0 &  & 3               & 1             \\
\,\;\textbf{+ CVNet-Rerank (Rerank Top-400)}              & \textbf{87.9}  & \textbf{80.7} & \textbf{90.5} & \textbf{82.4} &           & \textbf{75.6} & \textbf{65.1} & \textbf{80.2} & \textbf{67.3} &  & 3               & 1             \\
\textbf{R101-CVNet-Global (GLDv2-clean)}    & 80.2  & 74.0 & 90.3 & 80.6 &           & 63.1 & 53.7 & 79.1 & 62.2 &  & 3               & -             \\
\,\;\textbf{+ CVNet-Rerank (Rerank Top-100)}              & 85.6  & 79.6 & 90.6 & 81.5 &           & 72.9 & 64.5 & 80.4 & 66.2 &  & 3               & 1             \\
\,\;\textbf{+ CVNet-Rerank (Rerank Top-200)}              & 86.4  & 81.0 & 91.1 & 82.7 &           & 74.6 & 66.6 & 81.0 & 68.0 &  & 3               & 1             \\
\,\;\textbf{+ CVNet-Rerank (Rerank Top-400)}              & \textbf{87.2}  & \textbf{81.9} & \textbf{91.2} & \textbf{83.8} &           & \textbf{75.9} & \textbf{67.4} & \textbf{81.1} & \textbf{69.3} &  & 3               & 1             \\ \bottomrule
\end{tabular}
}
\vspace{-0.2cm}
\caption{\textbf{Comparison with state-of-the-art methods.} Performance comparison on $\mathcal{R}$Oxf/$\mathcal{R}$Par and 1M-added experiments (referred to as +1M) with Medium and Hard evaluation protocols. The proposed image retrieval framework outperforms state-of-the-art image retrieval methods by a large margin for every measure. The best and second-best scores are presented as \textbf{boldfaced} and \uline{underlined} text, respectively.
}
\label{tbl:main}
\vspace{-0.5cm}
\end{table*}

\vspace{-0.45cm}
\paragraph{Curriculum learning.}
To prevent hard samples from interfering with early learning, we apply hard negative mining and Hide-and-Seek in a curriculum learning manner. Instead of focusing on hard negatives from the outset, the rate of selecting hard negatives $r_{H}$ and the probability of Hide-and-Seek augmentation $p_{has}$ gradually increase as learning progresses.
This curriculum learning helps the network to retain its generality to ensure that it consistently performs well even when the re-ranking range is extended.

\vspace{-0.15cm}
\section{Experiments}
\label{sec:5.Experiments}

\vspace{-0.15cm}
\subsection{Implementation Details}
\label{sec:5.1.Implementation_Details}
\vspace{-0.15cm}
\paragraph{Common setting.}
Our proposed CVNet is implemented using PyTorch \cite{paszke2019pytorch}. We use the `clean' subset \cite{yokoo2020two} of Google Landmarks dataset v2 (1.58M images from 81k landmarks) \cite{weyand2020google} as a training set. The input image is augmented with random cropping/aspect ratio distortion and resized to $512 \times 512$. We use an SGD optimizer with a momentum of 0.9 and use cosine learning rate scheduling.

\vspace{-0.45cm}
\paragraph{Global backbone network.}
We use ResNet-50 (R50) and ResNet-101 (R101) as the encoder of global backbone networks with ImageNet \cite{russakovsky2015imagenet} pre-trained weights, whereas ResNet-50 is used for ablation studies.
We use a Shuffling Batch Normalization \cite{he2020momentum}, global descriptor size of 2048, and a queue size of 73,728. We set the $\tau$ to $1/30$, $m$ to $0.15$, $\eta$ to 0.999, and $\lambda_{cls}$ and $\lambda_{con}$ to $0.5$. The global model is trained for 25 epochs (39.5M steps) for the training dataset, using a learning rate of 0.005625, and a batch size of 144.

\vspace{-0.45cm}
\paragraph{Re-ranking network.}
For cross-scale correlation construction, we use $S=3$ scales (\ie $\{1/2, 1/\sqrt{2}, 1\}$). We extract the feature map $\mathbf{F}$ from the $f_3$ output and compress its channel dimension to $C'_l=256$.
Our training set contains various views of landmarks, including cases with no overlap. To avoid query-positive non-overlapping, we select verified match pairs for each class with help of deep local features \cite{noh2017large} and exclude only those classes with a number of verified match pairs. 
Please see the supplementary material for a more detailed explanation of the data selection and sampling process used for the CVNet-Rerank. Finally, we select 1M images from 31k landmarks, and the proposed re-ranking model is trained for 200 epochs (6.3M steps) for all classes, using a learning rate of 0.00375 and a batch size of 96.
$r_H$ and $p_{has}$ linearly increase from 0.2 to 1.0 and from 0 to 0.2 while training, respectively. 

\vspace{-0.45cm}
\paragraph{Feature extraction and matching.}
For global descriptor extraction, we follow the convention of previous studies \cite{gordo2017end,noh2017large,radenovic2018fine, cao2020unifying, tan2021instance}. We extract global descriptors of three scales: $\left\{1/\sqrt 2 ,1,\sqrt 2 \right\}$. The final global descriptor is calculated by L2-normalizing the average of the three descriptors.
During the re-ranking process, the final ranking is decided based on the final score $s_g + \alpha s_r$, where $s_g$ is the cosine similarity of the global descriptors, $s_r$ is the output score of the re-ranking network and $\alpha$ is the weight for $s_r$. As in previous studies \cite{cao2020unifying, revaud2019learning, teichmann2019detect, ng2020solar}, the weight $\alpha$ is tuned in $\mathcal{R}$Oxf/$\mathcal{R}$Par and fixed for its large-scale experiment and GLDv2-retrieval test. Finally, we set the $\alpha$ to 0.5.

\vspace{-0.15cm}
\subsection{Evaluation Benchmarks}
\label{sec:5.2.Evaluation_Benchmarks}
\vspace{-0.15cm}
We primarily evaluate our model on $\mathcal{R}$Oxford5k \cite{philbin2007object, radenovic2018revisiting} (referred to as $\mathcal{R}$Oxf) and $\mathcal{R}$Paris6k \cite{philbin2008lost, radenovic2018revisiting} (referred to as $\mathcal{R}$Par) datasets. Both datasets comprise 70 queries and 4933 and 6322 database images, respectively. In addition, an $\mathcal{R}$1M distractor set \cite{radenovic2018revisiting} is used for measuring the large-scale retrieval performance. Performance is measured using a mean Average Precision (mAP) metric.
Additionally, we evaluate our model on the instance-level large-scale image retrieval task of the Google Landmarks dataset v2 \cite{weyand2020google} (referred to as GLDv2-retrieval). The GLDv2-retrieval comprises 750 test query images and 762k database images. In this task, performance is evaluated using a mean Average Precision@100 (mAP$@100$) metric.

\vspace{-0.15cm}
\subsection{Results}
\label{sec:5.3.Results}
\vspace{-0.2cm}
In this section, we compare our model with state-of-the-art image retrieval methods.
\vspace{-0.6cm}
\paragraph{Comparison with state-of-the-art methods. (\cref{tbl:main}, \cref{tab:gldv2_retrieval})}
\cref{tbl:main} shows a comparison between results of the proposed model and state-of-the-art image retrieval methods on $\mathcal{R}$Oxf and $\mathcal{R}$Par, and their +1M experiments. 
For all settings, the proposed CVNet outperforms the state-of-the-art methods. Our global model shows performance comparable to the state-of-the-art methods without additional modules, and our proposed re-ranking network exhibits superior performance without using expensive multi-scale inference. Because of the nature of re-ranking, the proposed model exhibits significantly superior performance in the difficult dataset ($\mathcal{R}$Oxf), for the difficult protocol (Hard), when a large number of images interfere (+1M). Our re-ranking method yields an improvement of up to 14.9$\%$ (R50-$\mathcal{R}$Oxf-Hard+1M), which is significantly higher than any of the state-of-the-art methods. In addition, the proposed method performs well without loss of generality even when the number of re-ranking samples increases. \cref{tab:gldv2_retrieval} compares CVNet with the results of the previous study's GLDv2-retrieval test. Even in this comparison, our proposed CVNet outperforms all state-of-the-art methods.

\begin{table}[t]
\centering
\small
\resizebox{0.9\linewidth}{!}
{
\begin{tabular}{lc}
\toprule
\multicolumn{1}{c}{Method}                          & mAP@100       \\ \midrule
DELF-R-ASMK*+SP \cite{teichmann2019detect}          & 18.8          \\
R101-GeM+ArcFace \cite{weyand2020google}            & 20.7          \\
R101-GeM+CosFace \cite{yokoo2020two}                & 21.4          \\
R50-DELG (GLDv2-clean) \cite{cao2020unifying}       & 24.1          \\
\,\;+ GV (Rerank Top-100) \cite{cao2020unifying}    & 24.3          \\
R101-DELG (GLDv2-clean) \cite{cao2020unifying}      & 26.0          \\
\,\;+ GV (Rerank Top-100) \cite{cao2020unifying}    & 26.8          \\
\textbf{R50-CVNet-Global (Ours)}                    & 30.2          \\
\,\;\textbf{+ CVNet-Rerank (Rerank Top-100) (Ours)} & \textbf{32.4} \\
\textbf{R101-CVNet-Global (Ours)}                   & 32.5          \\
\,\;\textbf{+ CVNet-Rerank (Rerank Top-100) (Ours)} & \textbf{34.9} \\ \bottomrule
\end{tabular}
}
\vspace{-0.2cm}
\caption{\textbf{GLDv2-retrieval evaluation.} The result on the test split of the GLDv2-retrieval. The best scores are presented as \textbf{boldfaced} text for each ResNet backbone.
}
\label{tab:gldv2_retrieval}
\vspace{-0.5cm}
\end{table}

\vspace{-0.25cm}
\paragraph{Comparison with other re-ranking methods. (\cref{tbl:Comparison_ohter_reranking_methods})}
For a fair comparison, we attach the local branch of the DELG \cite{cao2020unifying} to our global backbone to learn the local DELG features. With these learned local features, we reproduce two re-ranking methods: geometric verification (GV) and Reranking Transformer \cite{tan2021instance}. Details of the reproduction are provided in the supplementary material.
While GV exhibits moderate performance improvement, RRT exhibits a decrease in performance in some sets, despite using the official code and setting. Our proposed method surpasses both methods by a large margin for all the measures.

\subsection{Ablation Experiments}
\label{sec:5.4.Ablation Experiments}
\vspace{-0.15cm}
In this section, we present the core ablation results in \cref{tbl:ablation_study}. Please refer to the supplementary material for a detailed explanation of this and additional ablation studies.

\vspace{-0.45cm}
\paragraph{Cross-scale correlation (\cref{tbl:Cross_Scale_Correlation}).}
We conduct an ablation study using cross-scale correlation construction to demonstrate its efficacy. The cross-scale correlation boosts the re-ranking performance, especially in hard protocols that include large-scale differences.

\vspace{-0.45cm}
\paragraph{Hard negative mining and Hide-and-Seek (\cref{tbl:Hard_Negative_Mining_and_Hide_and_Seek}).}
Our results demonstrate the effects of hard negative mining and Hide-and-Seek augmentation. When learning is performed only with random negatives, the network lost its distinguishing power and fails to re-rank. Considering the nature of re-ranking, that the process of re-ranking primarily encounters hard samples during testing, learning that focuses on hard negatives considerably improves performance. Hide-and-Seek augmentation also improves the overall performance by inducing the network to be robust against hard situations.

\vspace{-0.45cm}
\paragraph{Loss comparison for the CVNet-Global (\cref{tbl:Global_Loss_Comparison}).}
For the global backbone network, instead of using either the classification or contrastive loss, it is found that using both simultaneously results in overall improved performance.

\vspace{-0.45cm}
\paragraph{Quantization (\cref{tbl:8_bit_quantization}).}
To reduce the memory footprint, we conduct an experiment by quantizing the multi-scale features stored in 32 bits to 8 bits. While this quantization reduces the memory footprint by 1/4, it hardly diminishes the overall performance.

\vspace{-0.45cm}
\paragraph{Extraction latency and memory footprint (\cref{tbl:Extraction_Latency_and_Memory_Footprint}).}
Our feature extraction in the re-ranking process requires only a single inference, which is included in the process of extracting the global descriptor. Therefore, it has the lowest extraction latency time among the reproduced re-ranking methods. The memory footprint of the original model is large because of its dense nature. Thus, we attempt to reduce it with quantization (CVNet$^Q$). Through channel reduction and quantization, we achieve a memory footprint similar to that of re-ranking methods using sparse features while significantly improving the performance. Latency and matching time are measured on NVIDIA TITAN RTX GPU and i5-9600K CPU, for squared images of side 512. The time measured in the CPU is marked with an $*$.

\begin{table}[t]
\centering
\Huge
\resizebox{1.0\linewidth}{!}
{%
\begin{tabular}{@{}clcccccccc@{}}
\toprule
\multirow{2}{*}{\#} & \multirow{2}{*}{Method} & \multicolumn{4}{c}{Medium} & \multicolumn{4}{c}{Hard} \\ \cmidrule(l){3-10} 
 &  & $\mathcal{R}$Oxf & +1M & $\mathcal{R}$Par & +1M & $\mathcal{R}$Oxf & +1M & $\mathcal{R}$Par & +1M \\ \midrule
0 & CVNet-Global & 81.0 & 72.6 & 88.8 & 79.0 & 62.1 & 50.2 & 76.5 & 60.2 \\ \midrule
\multirow{3}{*}{100} & GV$^\dagger$ \cite{cao2020unifying} & {\ul 82.2} & {\ul 74.0} & {\ul 89.0} & {\ul 79.3} & 64.2 & 51.9 & {\ul 77.1} & {\ul 60.8} \\
 & RRT$^\dagger$ \cite{tan2021instance} & {\ul 82.2} & 72.4 & 88.8 & 78.8 & {\ul 66.1} & {\ul 52.3} & 75.6 & 57.4 \\
 & CVNet-Rerank & \textbf{86.1} & \textbf{77.6} & \textbf{89.4} & \textbf{79.9} & \textbf{72.8} & \textbf{61.1} & \textbf{78.6} & \textbf{63.9} \\ \midrule
\multirow{3}{*}{200} & GV$^\dagger$ \cite{cao2020unifying} & {\ul 82.7} & {\ul 74.8} & {\ul 89.1} & {\ul 79.4} & 65.0 & {\ul 52.3} & {\ul 77.5} & {\ul 60.8} \\
 & RRT$^\dagger$ \cite{tan2021instance} & 82.1 & 71.6 & 88.7 & 77.9 & {\ul 66.0} & 51.3 & 75.2 & 53.5 \\
 & CVNet-Rerank & \textbf{87.2} & \textbf{78.9} & \textbf{90.0} & \textbf{81.2} & \textbf{74.5} & \textbf{62.9} & \textbf{79.5} & \textbf{66.0} \\ \midrule
\multirow{3}{*}{400} & GV$^\dagger$ \cite{cao2020unifying} & {\ul 82.5} & {\ul 74.8} & {\ul 89.1} & {\ul 79.5} & 63.8 & {\ul 52.1} & {\ul 77.5} & {\ul 61.1} \\
 & RRT$^\dagger$ \cite{tan2021instance} & 81.7          & 71.2          & 88.2          & 75.2          & {\ul 65.2}    & 50.4          & 74.8          & 49.9 \\
 & CVNet-Rerank  & \textbf{87.9} & \textbf{80.7} & \textbf{90.5} & \textbf{82.4} & \textbf{75.6} & \textbf{65.1} & \textbf{80.2} & \textbf{67.3} \\ \bottomrule
\end{tabular}
}
\vspace{-0.2cm}
\caption{\textbf{Comparison with other re-ranking methods.} Geometric Verification (GV) and Reranking Transformers (RRT) are reproduced based on our R50-CVNet-Global.  $\dagger$ indicates reproduced. \# is the number of samples that is re-ranked and the best and second-best scores are presented as \textbf{boldfaced} and \uline{underlined} text, respectively.}
\label{tbl:Comparison_ohter_reranking_methods}
\vspace{-0.5cm}
\end{table}

\begin{table*}[ht]
\centering

\begin{subtable}{0.5\linewidth}
{
\centering

\resizebox{0.95\columnwidth}{!}
{
\Huge
\begin{tabular}{cccccccccc}
\toprule
\multirow{2}{*}{\#} & \multirow{2}{*}{CSC} & \multicolumn{4}{c}{Medium} & \multicolumn{4}{c}{Hard} \\ \cmidrule(l){3-10} 
 &  & $\mathcal{R}$Oxf & +1M & $\mathcal{R}$Par & +1M & $\mathcal{R}$Oxf & +1M & $\mathcal{R}$Par & +1M \\ \midrule
0 &  & 81.0 & 72.6 & 88.8 & 79.0 & 62.1 & 50.2 & 76.5 & 60.2 \\ \midrule
\multirow{2}{*}{100} &  & 84.9 & 76.1 & 88.8 & 79.3 & 69.9 & 57.4 & 76.3 & 61.1 \\
 & \checkmark & \textbf{86.1} & \textbf{77.6} & \textbf{89.4} & \textbf{79.9} & \textbf{72.8} & \textbf{61.1} & \textbf{78.6} & \textbf{63.9} \\ \midrule
\multirow{2}{*}{200} &  & 85.3 & 76.7 & 88.9 & 79.5 & 70.5 & 58.3 & 76.3 & 61.5 \\
 & \checkmark & \textbf{87.2} & \textbf{78.9} & \textbf{90.0} & \textbf{81.2} & \textbf{74.5} & \textbf{62.9} & \textbf{79.5} & \textbf{66.0} \\ \midrule
\multirow{2}{*}{400} &  & 85.5 & 77.6 & 89.0 & 79.7 & 70.7 & 59.3 & 76.4 & 61.6 \\
 & \checkmark & \textbf{87.9} & \textbf{80.7} & \textbf{90.5} & \textbf{82.4} & \textbf{75.6} & \textbf{65.1} & \textbf{80.2} & \textbf{67.3} \\ \bottomrule
\end{tabular}
}
\caption{\textbf{Cross-Scale Correlation.}}
\label{tbl:Cross_Scale_Correlation}


\resizebox{0.95\columnwidth}{!}
{
\Huge
\begin{tabular}{ccccccccccc}
\toprule
\multirow{2}{*}{\#} & \multirow{2}{*}{HNM} & \multirow{2}{*}{HaS} & \multicolumn{4}{c}{Medium} & \multicolumn{4}{c}{Hard} \\ \cmidrule(l){4-11} 
 &  & & $\mathcal{R}$Oxf & +1M & $\mathcal{R}$Par & +1M & $\mathcal{R}$Oxf & +1M & $\mathcal{R}$Par & +1M \\ \midrule
0 &  & & 81.0 & 72.6 & 88.8 & 79.0 & 62.1 & 50.2 & 76.5 & 60.2 \\ \midrule
\multirow{3}{*}{100} &  &  & 81.4 & 72.7 & 88.8 & 79.0 & 62.4 & 50.3 & 76.4 & 60.2 \\
 & \checkmark & & 85.8 & 77.5 & 89.3 & 79.9 & 71.6 & 60.5 & 78.1 & 63.7 \\
 & \checkmark & \checkmark & \textbf{86.1} & \textbf{77.6} & \textbf{89.4} & \textbf{79.9} & \textbf{72.8} & \textbf{61.1} & \textbf{78.6} & \textbf{63.9} \\ \midrule
\multirow{3}{*}{200} &  & & 81.3 & 72.6 & 88.7 & 78.9 & 62.5 & 50.2 & 76.5 & 60.2 \\
 & \checkmark & & 86.9 & 78.7 & 89.7 & 81.0 & 73.4 & 62.1 & 78.6 & 65.6 \\
 & \checkmark & \checkmark & \textbf{87.2} & \textbf{78.9} & \textbf{90.0} & \textbf{81.2} & \textbf{74.5} & \textbf{62.9} & \textbf{79.5} & \textbf{66.0}\\ \midrule
\multirow{3}{*}{400} &  & & 81.2 & 72.5 & 88.8 & 78.9 & 62.5 & 50.2 & 76.9 & 60.4  \\
 & \checkmark & & 87.5 &80.3&89.9&82.0&74.2&64.3&78.9&66.4 \\
 & \checkmark & \checkmark & \textbf{87.9} & \textbf{80.7} & \textbf{90.5} & \textbf{82.4} & \textbf{75.6} & \textbf{65.1} & \textbf{80.2} & \textbf{67.3} \\ \bottomrule
\end{tabular}
}
\caption{\textbf{Hard Negative Mining (HNM) and Hide-and-Seek (HaS).}}
\label{tbl:Hard_Negative_Mining_and_Hide_and_Seek}
}
\end{subtable}%
\hfill
\begin{subtable}{0.5\linewidth}
{
\centering

\resizebox{0.88\columnwidth}{!}
{
\Huge
\begin{tabular}{cccccccccc}
\toprule
\multirow{2}{*}{$\mathcal{L}_{cls}$} & \multirow{2}{*}{$\mathcal{L}_{con}$} & \multicolumn{4}{c}{Medium} & \multicolumn{4}{c}{Hard} \\ \cmidrule(l){3-10} 
           &            & $\mathcal{R}$Oxf & +1M & $\mathcal{R}$Par & +1M & $\mathcal{R}$Oxf & +1M & $\mathcal{R}$Par & +1M \\ \midrule
\checkmark &            & {\ul 78.0} & 69.4 & \textbf{89.8} & {\ul 77.3} & 57.1 & 42.9 & \textbf{78.4} & {\ul 56.9} \\
           & \checkmark & 80.1 & \textbf{73.5} & 87.7 & 76.2 & \textbf{62.2} & \textbf{51.9} & 74.0 & 56.4 \\
\checkmark & \checkmark & \textbf{81.0} & {\ul 72.6} & {\ul 88.8} & \textbf{79.0} & {\ul 62.1} & {\ul 50.2} & {\ul 76.5} & \textbf{60.2} \\ \bottomrule
\end{tabular}
}
\caption{\textbf{Loss Comparison of CVNet-Global.}}
\label{tbl:Global_Loss_Comparison}

\resizebox{0.88\columnwidth}{!}
{
\Huge
\begin{tabular}{cccccccccc}
\toprule
\multirow{2}{*}{\#} & \multirow{2}{*}{\begin{tabular}[c]{@{}c@{}}8-bit\\quant\end{tabular}} & \multicolumn{4}{c}{Medium} & \multicolumn{4}{c}{Hard} \\ \cmidrule(l){3-10} 
 &  & $\mathcal{R}$Oxf & +1M & $\mathcal{R}$Par & +1M & $\mathcal{R}$Oxf & +1M & $\mathcal{R}$Par & +1M \\ \midrule
0 &  & 81.0 & 72.6 & 88.8 & 79.0 & 62.1 & 50.2 & 76.5 & 60.2 \\ \midrule
\multirow{2}{*}{100} &  & \textbf{86.1} & \textbf{77.6} & \textbf{89.4} & \textbf{79.9} & \textbf{72.8} & \textbf{61.1} & \textbf{78.6} & \textbf{63.9} \\
 & \checkmark       &   \textbf{86.1} & \textbf{77.6} & \textbf{89.4} & \textbf{79.9} & \textbf{72.8} & \textbf{61.1} & \textbf{78.6} & \textbf{63.9} \\ \midrule
\multirow{2}{*}{200} &  & \textbf{87.2} & \textbf{78.9} & \textbf{90.0} & \textbf{81.2} & \textbf{74.5} & \textbf{62.9} & \textbf{79.5} & \textbf{66.0} \\
 & \checkmark        &  \textbf{87.2} & \textbf{78.9} & \textbf{90.0} & \textbf{81.2} & \textbf{74.5} & 62.8 & \textbf{79.5} & \textbf{66.0} \\ \midrule
\multirow{2}{*}{400} &  & \textbf{87.9} & \textbf{80.7} & \textbf{90.5} & \textbf{82.4} & \textbf{75.6} & \textbf{65.1} & \textbf{80.2} & \textbf{67.3} \\
 & \checkmark           & \textbf{87.9} & 80.6 & \textbf{90.5} & \textbf{82.4} & 75.5 & \textbf{65.1} & \textbf{80.2} & \textbf{67.3} \\ \bottomrule
\end{tabular}
}
\caption{\textbf{8-bit Quantization.}}
\label{tbl:8_bit_quantization}


\resizebox{0.88\columnwidth}{!}
{
\Huge
\begin{tabular}{@{}llccccccclcc@{}}
\toprule
\multicolumn{1}{c}{\multirow{2}{*}{Method}} &  & \multicolumn{2}{c}{Multi-scale} &  & \multicolumn{3}{c}{\begin{tabular}[c]{@{}c@{}}Extraction\\ latency (ms)\end{tabular}} & \begin{tabular}[c]{@{}c@{}}Matching\\ time (ms)\end{tabular} &  & \multicolumn{2}{c}{\begin{tabular}[c]{@{}c@{}}Memory\\ (GB)\end{tabular}} \\ \cmidrule(l){2-12} 
\multicolumn{1}{c}{} &  & global & local &  & global & +local & total  &  &  & $\mathcal{R}$Oxf & $\mathcal{R}$Par \\ \midrule
DELG$^\dagger$ &  & 3 & 7 &  & 24.0 & 33.1 & 57.1 & 69.0$^*$ &  & 4.25 & 5.35 \\
RRT$^\dagger$ &  & 3 & 7 &  & 24.0 & 33.1 & 57.1 & \textbf{3.2} &  & \textbf{2.16} & \textbf{2.72} \\
CVNet &  & 3 & \textbf{1} &  & 24.0 & \textbf{1.7} & \textbf{25.7} & 15.6 &  & 27.02 & 33.55 \\
CVNet$^Q$ &  & 3 & \textbf{1} &  & 24.0 & \textbf{1.7} & \textbf{25.7} & 15.6 &  & 6.88 & 8.52 \\ \bottomrule
\end{tabular}
}

\caption{\textbf{Extraction Latency and Memory Footprint.}}
\label{tbl:Extraction_Latency_and_Memory_Footprint}

}
\end{subtable}
\vspace{-0.2cm}
\caption{\textbf{Ablation study for CVNet.} mAP measures for each setting. \# is the number of samples that are re-ranked.}
\label{tbl:ablation_study}
\end{table*}
\begin{figure*}[t]
\vspace{-0.25cm}
\centering
\includegraphics[width=0.9\linewidth]{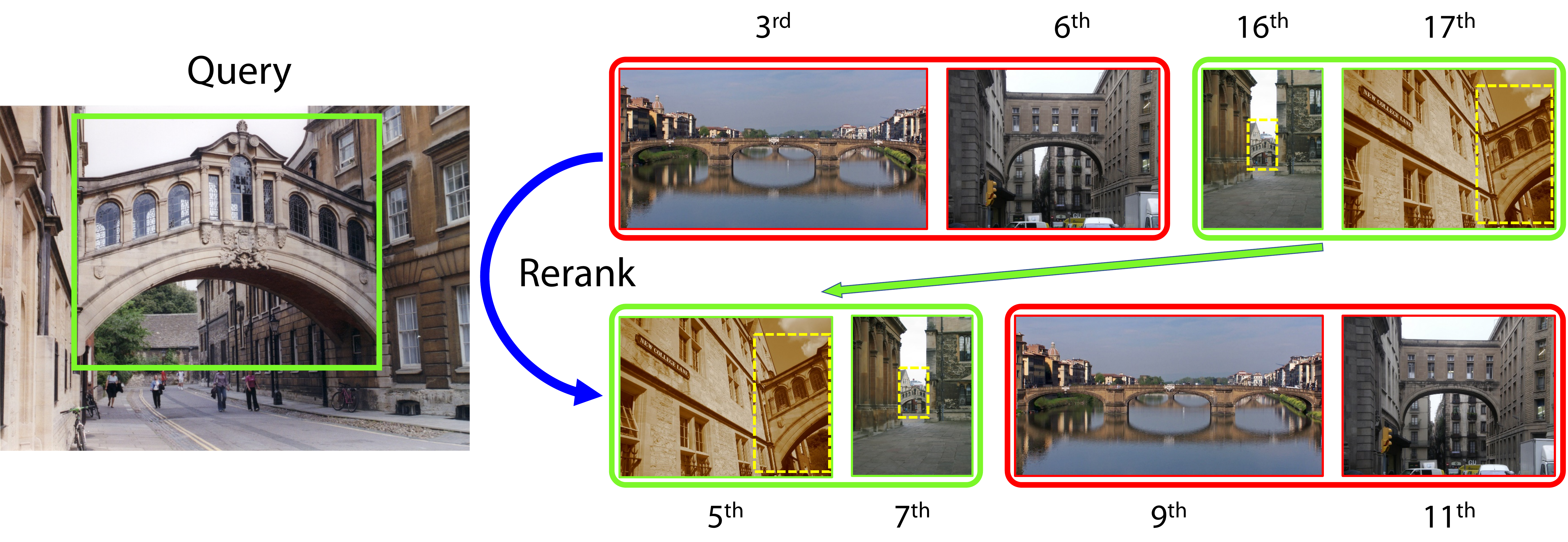}
\vspace{-0.2cm}
\caption{
Example qualitative results on $\mathcal{R}$Oxf-Hard+1M with R50-CVNet. The upper row shows the global descriptor matching result and the lower row shows the re-ranking result. Correct/incorrect results are marked with \textcolor{green}{green}/\textcolor{red}{red} borders, respectively. The query used as an input is generated by cropping only the part bounded by a green square. A dashed \textcolor{yellow}{yellow} line indicates the areas that overlap with the query.
}
\vspace{-0.5cm}
\label{fig:fig7.qualitative_results}
\end{figure*}    

\vspace{-0.3cm}
\section{Discussion}
\label{sec:6.Discussion}
\vspace{-0.15cm}

\paragraph{Qualitative results.}
Examples of our re-ranking results are provided in \cref{fig:fig7.qualitative_results}.  Despite technological advances, global descriptor matching is easily fooled by similar-looking negative images and has difficulty finding occluded or truncated positives, even more so at different scales.
Our re-ranking network can respond to scale changes owing to cross-scale correlation and has been trained to be robust in situations involving challenges such as occlusion. Consequently, our re-ranking network shows robust final retrieval results by boosting the ranks of positives even in cases where global descriptors are misjudged.
Additional qualitative results are provided in the supplementary material.

\vspace{-0.45cm}
\paragraph{Limitations and future work.}
Although our proposed re-ranking method has significant potential, it has shortcomings in terms of speed and memory, owing to its dense nature. To solve this problem, we apply kernel sparsification, channel reduction, and quantization to bring them up to an appropriate level, but the proposed method still requires considerable improvement. Our future work will aim to achieve improvements in speed and memory while preserving its strong performance.

\vspace{-0.15cm}
\section{Conclusion}
\label{sec:7.Conclusion}
\vspace{-0.15cm}
In this study, we propose a novel image retrieval re-ranking network that directly predicts similarity by leveraging dense feature correlation in a convolutional manner.
We design the network to construct cross-scale correlations within a single inference, thereby enabling cross-scale matching instead of expensive multi-scale inferences.
Considering that re-ranking primarily encounters hard samples during testing, we trained this network by focusing on hard samples.
With the aforementioned contributions, we achieve state-of-the-art performance on several benchmarks, demonstrating that dense feature correlation is a powerful cue for image retrieval re-ranking.

\vspace{-0.4cm}
\paragraph{Acknowledgements.}
This work was supported by the Industry Core Technology Development Project, 20005062, Development of Artificial Intelligence Robot Autonomous Navigation Technology for Agile Movement in Crowded Space, funded by the Ministry of Trade, industry \& Energy (MOTIE, Republic of Korea).

{\small
\bibliographystyle{ieee_fullname}
\bibliography{main}
}

\clearpage


\section*{Supplementary material}

# Supplementary Material:
# Correlation Verification for Image Retrieval

Seongwon Lee      Hongje Seong      Suhyeon Lee      Euntai Kim*

School of Electrical and Electronic Engineering, Yonsei University, Seoul, Korea

{won4113, hjseong, hyeon93, etkim}@yonsei.ac.kr

## S1. Data Selection and Sampling Process

**Overlapped positive selection.** In this study, we use the 'clean' subset [18] of Google Landmarks dataset v2 (1.58M images from 81k landmarks) [16] as a training set. This dataset has large intra-class variability and includes multiple viewpoints, such as indoor and outdoor views of landmarks. Therefore, when sampling the same-class image pair from this dataset, we cannot guarantee an overlap between the two images, and non-overlapping query-positive pairs can interfere with learning image matching. To avoid the non-overlapping case, we select overlapped pairs for each class in advance with the help of the DELF [9] local feature. The overall process is similar to the data cleaning process of [18]: The primary difference is that [18] aims to remove outlier data from the dataset, whereas we aim to select same-class pairs that actually overlap. To select an overlapped pair, for every dataset sample $x_i$, we first select up to ten of the nearest neighbors that are assigned to the same class as $x_i$ with a global descriptor extracted from R50-CVNet-Global. After the nearest neighbors are selected, spatial verification using RANSAC with a pre-trained DELF feature is performed on the nearest neighbors selected for each sample. Subsequently, we select the pair with 30 or more inlier matches as an overlapped pair. Furthermore, only classes with more than 10 samples belonging to overlapped pairs are used for training. Finally, we select 1M images from 31k landmarks of the GLDv2-clean dataset and use this subset as a training set for CVNet-Rerank. Although this selection process is quite expensive because of the use of RANSAC, it only needs to be performed once.

**Sampling process.** CVNet-Rerank is trained for 200 epochs (6.3M steps) for all selected classes. For every epoch, we construct tuples of query, positive, and negative samples for each class. The query image is randomly sampled from each class, a positive image is randomly chosen from among the overlapped positives of the query, and a negative image is sampled from random or hard-negative

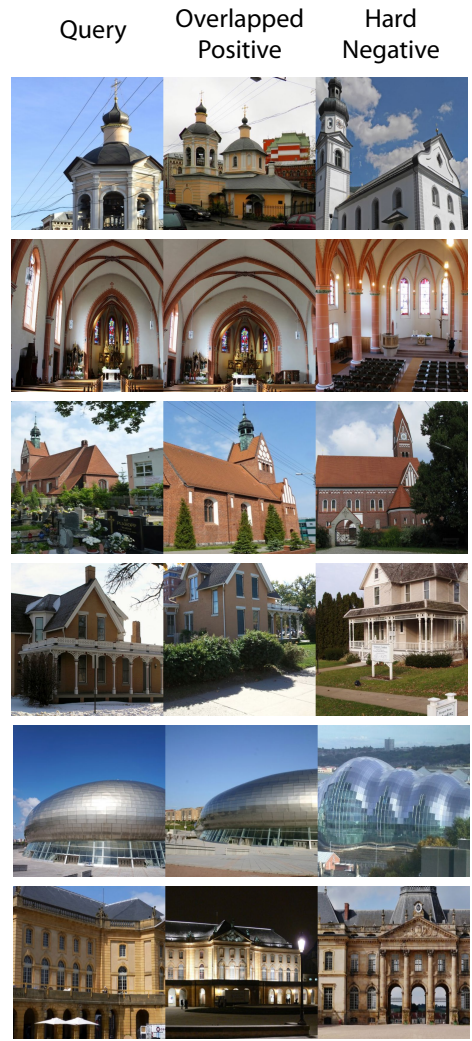

Figure S1. Example of query, overlapped positive, and hard negative samples sampled from the selected subset of the GLDv2-clean dataset. Our proposed re-ranking network learns better discrimination ability by learning the cue for equivalence from a overlapped positive and the cue for the difference from a hard negative.

---
*Corresponding author.

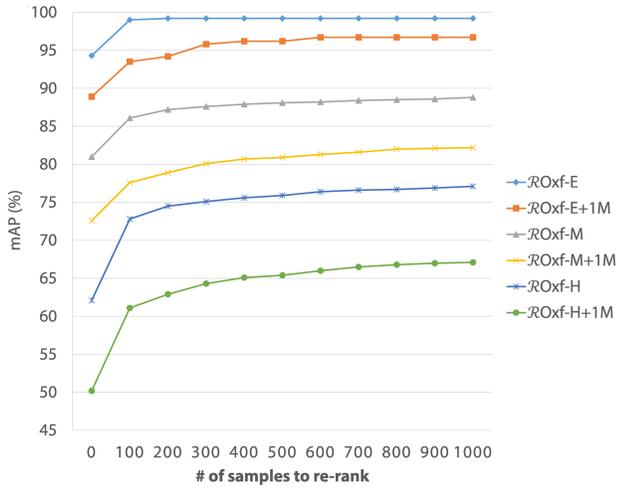
(a) ROxford5k and its +1M Experiments.

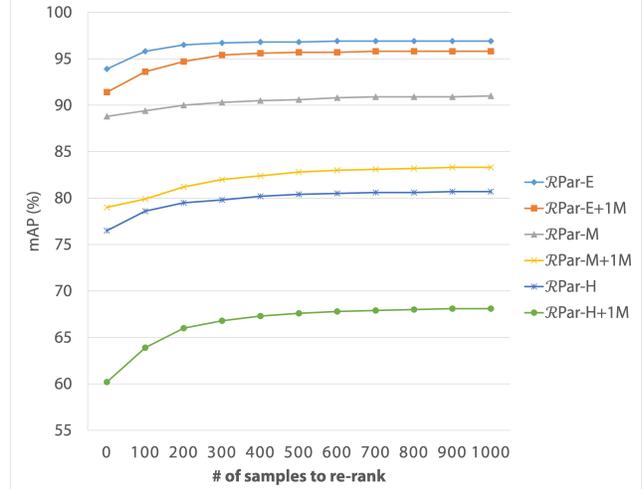
(b) RParis6k and its +1M Experiments.

Figure S2. **Analysis about number of samples to re-rank.**

| rerank | $r_H$ | Medium | | | | Hard | | | |
|---|---|---|---|---|---|---|---|---|---|
| | | ROxf | +1M | RPar | +1M | ROxf | +1M | RPar | +1M |
| 0 | | 81.0 | 72.6 | 88.8 | 79.0 | 62.1 | 50.2 | 76.5 | 60.2 |
| 100 | 0.0 | 81.6 | 72.8 | 88.8 | 79.0 | 62.6 | 50.2 | 76.6 | 60.3 |
| | 1.0 | 85.5 | 77.1 | 89.3 | **80.0** | 70.8 | 59.5 | 77.5 | 63.7 |
| | 0.5 | 85.5 | 77.3 | 89.2 | 79.8 | 71.5 | 60.5 | 77.9 | 63.4 |
| | 0.2-1.0 | **86.1** | **77.6** | **89.4** | 79.9 | **72.8** | **61.1** | **78.6** | **63.9** |
| 200 | 0.0 | 81.5 | 72.7 | 88.8 | 79.0 | 62.6 | 50.1 | 76.7 | 60.4 |
| | 1.0 | 86.2 | 78.2 | 89.5 | 81.1 | 71.5 | 60.7 | 76.5 | 65.3 |
| | 0.5 | 86.4 | 78.4 | 89.7 | 80.8 | 73.1 | 62.0 | 78.7 | 65.3 |
| | 0.2-1.0 | **87.2** | **78.9** | **90.0** | **81.2** | **74.5** | **62.9** | **79.5** | **66.0** |
| 400 | 0.0 | 81.4 | 72.6 | 88.9 | 79.1 | 62.6 | 50.1 | 77.1 | 60.6 |
| | 1.0 | 86.2 | 79.5 | 88.5 | 81.8 | 71.0 | 62.0 | 74.0 | 65.5 |
| | 0.5 | 86.9 | 79.8 | 90.3 | 82.0 | 74.0 | 63.9 | 79.5 | 66.7 |
| | 0.2-1.0 | **87.9** | **80.7** | **90.5** | **82.4** | **75.6** | **65.1** | **80.2** | **67.3** |

Table S1. **Hard-Negative Sampling Ratio.**

| rerank | $p_{has}$ | Medium | | | | Hard | | | |
|---|---|---|---|---|---|---|---|---|---|
| | | ROxf | +1M | RPar | +1M | ROxf | +1M | RPar | +1M |
| 0 | | 81.0 | 72.6 | 88.8 | 79.0 | 62.1 | 50.2 | 76.5 | 60.2 |
| 100 | 0.0 | 85.8 | 77.5 | 89.3 | 79.9 | 71.6 | 60.5 | 78.1 | 63.7 |
| | 0.2 | 86.1 | 77.1 | 89.3 | 79.9 | 72.3 | 60.1 | 78.1 | 63.6 |
| | 0.0-0.2 | **86.1** | **77.6** | **89.4** | 79.9 | **72.8** | **61.1** | **78.6** | **63.9** |
| 200 | 0.0 | 86.9 | 78.7 | 89.7 | 81.0 | 73.4 | 62.1 | 78.6 | 65.6 |
| | 0.2 | 87.1 | 78.3 | 89.7 | 81.1 | 74.0 | 61.8 | 78.7 | 65.7 |
| | 0.0-0.2 | **87.2** | **78.9** | **90.0** | **81.2** | **74.5** | **62.9** | **79.5** | **66.0** |
| 400 | 0.0 | 87.5 | 80.3 | 89.9 | 82.0 | 74.2 | 64.3 | 78.9 | 66.4 |
| | 0.2 | 87.8 | 80.1 | 90.1 | 82.2 | 75.1 | 64.0 | 79.3 | 66.8 |
| | 0.0-0.2 | **87.9** | **80.7** | **90.5** | **82.4** | **75.6** | **65.1** | **80.2** | **67.3** |

Table S2. **Hide-and-Seek Probability.**

samples according to the hard negative sampling ratio $r_{neg}$. Fig. S1 shows examples of our sampling results. By learning with well-constructed training pairs, the network can achieve improved discriminating ability.

## S2. Additional Ablation Studies and Analysis
### S2.1. Curriculum Learning

Learning focused on hard samples can improve the robustness of the network in hard situations. However, this could lead to a loss of generality. Accordingly, we apply curriculum learning to focus on hard samples without losing generality. In this subsection, we show that the proposed network performs re-ranking well regardless of the matching difficulty with the help of curriculum learning. Furthermore, we show a more detailed analysis of curriculum learning.

**Generality of learning (Fig. S2).** By gradually increasing the number of samples to be re-ranked, we can verify whether the network distinguishes hard samples well while retaining generality for normal samples. As shown in Fig. S2, the proposed re-ranking network dramatically improves performance when it is applied to top ranks where many hard samples exist. Even if the re-ranking targets are expanded to easier samples, our proposed re-ranking model continues to exhibit improved performance without losing generality.

**Hard negative mining (Tab. S1).** To prove the effectiveness of hard negative mining applied simultaneously with the curriculum approach, we conduct experiments by varying the hard-negative sampling ratio $r_H$. The results are presented in Tab. S1. When the network learns using randomly sampled negatives ($r_H = 0$), global retrieval results do not improve when re-ranking. This indicates that learning to discriminate hard samples using only random negative is difficult. Accordingly, when sampling hard negatives with a fixed ratio ($r_H = 1.0, 0.5$), the network exhibits a significantly improved performance. Moreover, when a hard-negative ratio is set through the curriculum manner ($r_H = 0.2\text{-}1.0$), the proposed re-ranking network exhibits its best performance. This proves that hard negatives are a critical key to re-ranking learning, and hard negative mining

| # | $C'_l$ | Medium | | | | Hard | | | |
|---|---|---|---|---|---|---|---|---|---|
| | | $\mathcal{R}$Oxf | +1M | $\mathcal{R}$Par | +1M | $\mathcal{R}$Oxf | +1M | $\mathcal{R}$Par | +1M |
| 0 | | 81.0 | 72.6 | 88.8 | 79.0 | 62.1 | 50.2 | 76.5 | 60.2 |
| 100 | 16 | 85.4 | 76.3 | 89.2 | 79.7 | 70.9 | 57.9 | 77.5 | 62.5 |
| | 32 | 85.5 | 76.5 | 89.2 | 79.8 | 71.7 | 59.7 | 77.6 | 63.1 |
| | 64 | 85.4 | 76.8 | 89.3 | 79.9 | 71.6 | 60.2 | 77.6 | 63.3 |
| | 128 | 85.5 | 76.9 | 89.3 | 79.9 | 71.4 | 60.0 | 77.9 | 63.6 |
| | 256 | **86.1** | **77.6** | **89.4** | **79.9** | **72.8** | **61.1** | **78.6** | **63.9** |
| | 512 | 85.5 | 76.9 | 89.3 | 79.9 | 71.3 | 59.7 | 77.8 | 63.7 |
| | 1024 | 85.7 | 77.3 | 89.4 | 80.0 | 71.6 | 59.7 | 78.1 | 63.7 |
| 200 | 16 | 86.2 | 77.2 | 89.6 | 80.5 | 71.9 | 58.8 | 78.1 | 63.9 |
| | 32 | 86.5 | 77.6 | 89.6 | 80.7 | 73.2 | 61.1 | 78.1 | 64.5 |
| | 64 | 86.2 | 77.9 | 89.7 | 80.9 | 72.9 | 61.3 | 77.9 | 64.9 |
| | 128 | 86.4 | 78.0 | 89.9 | 81.2 | 72.8 | 61.2 | 78.6 | 65.5 |
| | 256 | **87.2** | **78.9** | **90.0** | **81.2** | **74.5** | **62.9** | **79.5** | **66.0** |
| | 512 | 86.3 | 77.9 | 89.8 | 81.1 | 72.6 | 61.0 | 78.4 | 65.6 |
| | 1024 | 86.5 | 78.6 | 89.9 | 81.1 | 72.6 | 61.2 | 79.1 | 65.5 |
| 400 | 16 | 86.5 | 78.4 | 89.9 | 81.2 | 72.3 | 60.2 | 78.5 | 64.5 |
| | 32 | 87.0 | 79.1 | 89.7 | 81.5 | 74.1 | 63.0 | 78.2 | 65.1 |
| | 64 | 86.7 | 79.3 | 89.8 | 81.7 | 73.6 | 63.3 | 78.1 | 65.6 |
| | 128 | 87.0 | 79.6 | 90.1 | 82.1 | 73.5 | 63.2 | 78.9 | 66.3 |
| | 256 | **87.9** | **80.7** | **90.5** | **82.4** | **75.6** | **65.1** | **80.2** | **67.3** |
| | 512 | 86.6 | 79.3 | 90.0 | 82.0 | 73.1 | 62.8 | 78.5 | 66.3 |
| | 1024 | 87.0 | 80.1 | 90.3 | 82.1 | 73.4 | 63.0 | 79.6 | 66.6 |

Table S3. **Channel Compression.**

| # | type | Medium | | | | Hard | | | |
|---|---|---|---|---|---|---|---|---|---|
| | | $\mathcal{R}$Oxf | +1M | $\mathcal{R}$Par | +1M | $\mathcal{R}$Oxf | +1M | $\mathcal{R}$Par | +1M |
| 0 | | 81.0 | 72.6 | 88.8 | 79.0 | 62.1 | 50.2 | 76.5 | 60.2 |
| 100 | float32 | **86.1** | **77.6** | **89.4** | **79.9** | **72.8** | **61.1** | **78.6** | **63.9** |
| | int8 | 86.1 | 77.6 | 89.4 | 79.9 | 72.7 | 61.1 | 78.6 | 63.9 |
| | int4 | 86.0 | 77.3 | 89.3 | 79.8 | 72.5 | 60.5 | 78.0 | 63.5 |
| 200 | float32 | **87.2** | **78.9** | **90.0** | **81.2** | **74.5** | **62.9** | **79.5** | **66.0** |
| | int8 | 87.2 | 78.9 | 90.0 | 81.2 | 74.5 | 62.8 | 79.5 | 66.0 |
| | int4 | 86.9 | 78.6 | 89.7 | 81.0 | 73.8 | 62.1 | 78.7 | 65.3 |
| 400 | float32 | **87.9** | **80.7** | **90.5** | **82.4** | **75.6** | **65.1** | **80.2** | **67.3** |
| | int8 | 87.9 | 80.6 | 90.5 | 82.4 | 75.5 | 65.1 | 80.2 | 67.3 |
| | int4 | 87.4 | 80.1 | 90.1 | 82.0 | 74.6 | 63.8 | 79.3 | 66.3 |

Table S4. **Feature Quantization.**

| # | layer | Medium | | | | Hard | | | |
|---|---|---|---|---|---|---|---|---|---|
| | | $\mathcal{R}$Oxf | +1M | $\mathcal{R}$Par | +1M | $\mathcal{R}$Oxf | +1M | $\mathcal{R}$Par | +1M |
| 0 | | 81.0 | 72.6 | 88.8 | 79.0 | 62.1 | 50.2 | 76.5 | 60.2 |
| 100 | $f_3$ | **86.1** | **77.6** | **89.4** | **79.9** | **72.8** | **61.1** | **78.6** | **63.9** |
| | $f_4$ | 85.4 | 76.0 | 89.2 | 79.8 | 69.0 | 56.2 | 76.7 | 63.0 |
| | fuse | 85.2 | 76.8 | 89.3 | 79.9 | 69.6 | 57.6 | 77.0 | 63.2 |
| 200 | $f_3$ | **87.2** | **78.9** | **90.0** | **81.2** | **74.5** | **62.9** | **79.5** | **66.0** |
| | $f_4$ | 85.8 | 76.7 | 89.2 | 80.7 | 69.7 | 57.0 | 75.5 | 63.8 |
| | fuse | 85.4 | 77.5 | 89.4 | 80.9 | 69.8 | 58.2 | 75.8 | 64.0 |
| 400 | $f_3$ | **87.9** | **80.7** | **90.5** | **82.4** | **75.6** | **65.1** | **80.2** | **67.3** |
| | $f_4$ | 86.0 | 77.7 | 88.2 | 80.9 | 69.8 | 58.0 | 73.7 | 63.2 |
| | fuse | 85.1 | 78.3 | 88.2 | 80.9 | 68.9 | 58.3 | 74.0 | 63.3 |

Table S5. **Feature Extraction Layer Selection.**

is even more effective when used with curriculum learning.

**Hide-and-Seek (Tab. S2).** Similarly, to prove the effectiveness of the Hide-and-Seek [13] augmentation, we conduct experiments by varying the Hide-and-Seek probability $p_{has}$. Tab. S2 also shows that Hide-and-Seek is an appropriate strategy to help re-ranking learning and that it can be even more effective when used with curriculum learning.

### S2.2. Memory Footprint Reduction

Despite having significant potential, the proposed re-ranking method possesses a large memory owing to its dense nature. In this subsection, we present several effective solutions for reducing the memory footprint of the proposed re-ranking model.

**Channel compression (Tab. S3).** We pre-extract and store a multi-scale feature pyramid for every database sample for online re-ranking, which is where memory consumption primarily occurs. To reduce the memory footprint of the proposed model, we compress the channel of the feature map $C_l$ to $C'_l$ using a $3 \times 3$ convolution layer in the process of constructing the multi-scale feature pyramid. Here, we conduct experiments by varying the compressed channel dimension $C'_l$, to find a balance between memory footprint and re-ranking performance. The results are presented in Tab. S3. When the $C'_l$ is 256, the proposed re-ranking model exhibited its best performance; therefore, we finally selected $C'_l$ as 256 in our study. However, on systems where memory management is more important, choosing a smaller dimension, such as 16 ($\frac{1}{16}$ of our model's memory footprint) or 32 ($\frac{1}{8}$ of our model's memory footprint) can be a good option. This quantization reduces the memory footprint even further, albeit at the cost of a marginally reduced performance.

**Quantization (Tab. S4).** To reduce the memory burden, we measured the re-ranking performance while taking the correlation of quantized features as an input. The results are presented in Tab. S4. Similar to the case of channel compression, feature quantization also reduces the memory footprint at the risk of marginal performance degradation.

### S2.3. Model Design and Parameter Selection

In this subsection, we present several analyses of the design of the re-ranking model and its parameter selection.

**Feature extraction layer selection (Tab. S5).** First, we analyze CVNet-Global to determine which of its stages is more suited for use as an input for the re-ranking network. The results are presented in Tab. S5. $f_i$ denotes the $i$th *Res-Block*. When receiving an output of $f_4$ as an input, the stride and kernel size in the first block are reduced by 1 and 3, respectively; therefore, the output resolution is identical to that when an output of $f_3$ is received as an input. In the "fuse" case, both the output feature maps of $f_3$ and $f_4$ are received as input. In this case, the outputs of $f_3$ and $f_4$ pass through the first two convolutional blocks separately and

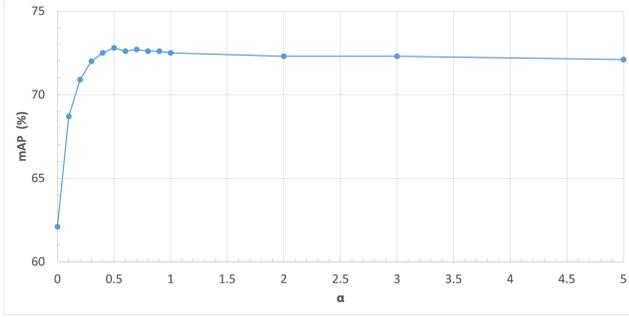
(a) $\mathcal{R}$Oxford5k-Hard Experiments.

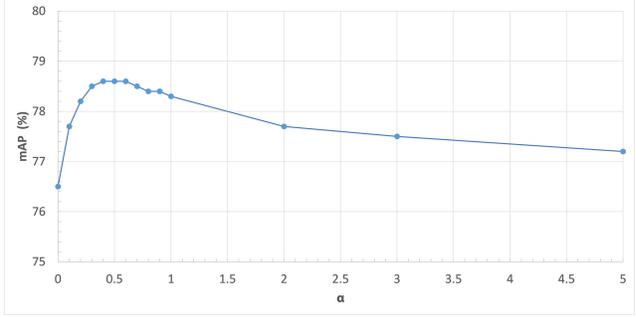
(b) $\mathcal{R}$Paris6k-Hard Experiments.

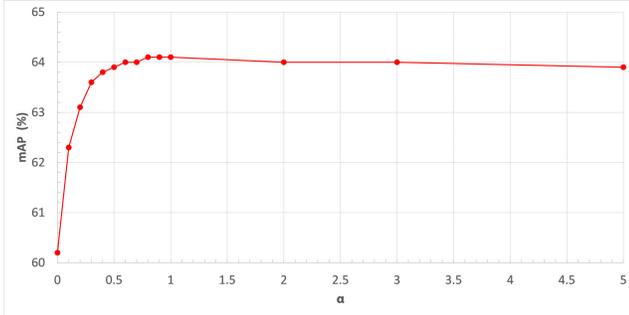
(c) $\mathcal{R}$Oxford5k-Hard+1M Experiments.

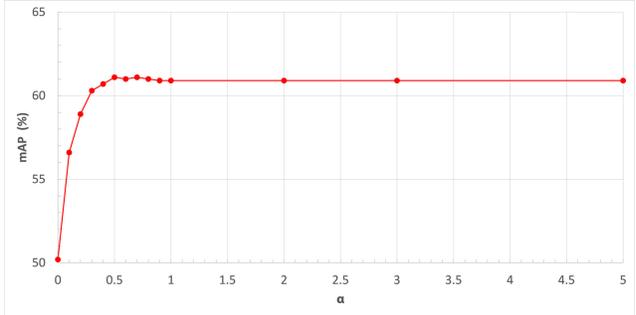
(d) $\mathcal{R}$Paris6k-Hard+1M Experiments.

Figure S3. **Experiments about the score fusion weight $\alpha$.** $\alpha$ is tuned in $\mathcal{R}$Oxf-Hard (Fig. S3a)/$\mathcal{R}$Par-Hard (Fig. S3b) and fixed for $\mathcal{R}$Oxf-Hard+1M (Fig. S3c)/$\mathcal{R}$Par-Hard+1M (Fig. S3d). We finally set an $\alpha$ to 0.5.

| # | Scale 1 | Scale $\frac{1}{2}$ | Scale $\frac{1}{\sqrt{2}}$ | Medium $\mathcal{R}$Oxf | +1M | $\mathcal{R}$Par | +1M | Hard $\mathcal{R}$Oxf | +1M | $\mathcal{R}$Par | +1M |
|---|---|---|---|---|---|---|---|---|---|---|---|
| 0 | | | | 81.0 | 72.6 | 88.8 | 79.0 | 62.1 | 50.2 | 76.5 | 60.2 |
| 100 | ✓ | | | 84.9 | 76.1 | 88.8 | 79.3 | 69.9 | 57.4 | 76.3 | 61.1 |
|  | ✓ | ✓ | | 85.8 | 77.1 | 89.3 | **80.0** | 71.5 | 59.8 | 78.3 | 63.9 |
|  | ✓ | | ✓ | 85.7 | 77.3 | 89.3 | 79.9 | 71.1 | 59.9 | 78.0 | 63.4 |
|  | ✓ | ✓ | ✓ | **86.1** | **77.6** | **89.4** | 79.9 | **72.8** | **61.1** | **78.6** | **63.9** |
| 200 | ✓ | | | 85.3 | 76.7 | 88.9 | 79.5 | 70.5 | 58.3 | 76.3 | 61.5 |
|  | ✓ | ✓ | | 86.6 | 78.2 | 90.0 | 81.2 | 72.8 | 61.3 | 79.0 | **66.0** |
|  | ✓ | | ✓ | 86.5 | 78.5 | 89.9 | 81.0 | 72.2 | 61.1 | 78.7 | 65.3 |
|  | ✓ | ✓ | ✓ | **87.2** | **78.9** | **90.0** | **81.2** | **74.5** | **62.9** | **79.5** | **66.0** |
| 400 | ✓ | | | 85.5 | 77.6 | 89.0 | 79.7 | 70.7 | 59.3 | 76.4 | 61.6 |
|  | ✓ | ✓ | | 87.1 | 79.9 | 90.3 | **82.4** | 73.6 | 63.4 | 79.3 | 67.2 |
|  | ✓ | | ✓ | 87.0 | 80.2 | 90.3 | 82.0 | 72.9 | 63.1 | 79.0 | 66.4 |
|  | ✓ | ✓ | ✓ | **87.9** | **80.7** | **90.5** | **82.4** | **75.6** | **65.1** | **80.2** | **67.3** |

Table S6. **Scale Selection.**

| # | kernel | Medium $\mathcal{R}$Oxf | +1M | $\mathcal{R}$Par | +1M | Hard $\mathcal{R}$Oxf | +1M | $\mathcal{R}$Par | +1M |
|---|---|---|---|---|---|---|---|---|---|
| 0 | | 81.0 | 72.6 | 88.8 | 79.0 | 62.1 | 50.2 | 76.5 | 60.2 |
| 100 | asymmetric | **86.1** | **77.6** | **89.4** | 79.9 | **72.8** | **61.1** | **78.6** | **63.9** |
|  | symmetric | 85.2 | 76.8 | 89.3 | 79.9 | 69.6 | 57.6 | 77.0 | 63.2 |
| 200 | asymmetric | **87.2** | **78.9** | **90.0** | **81.2** | **74.5** | **62.9** | **79.5** | **66.0** |
|  | symmetric | 85.4 | 77.5 | 89.4 | 80.9 | 69.8 | 58.2 | 75.8 | 64.0 |
| 400 | asymmetric | **87.9** | **80.7** | **90.5** | **82.4** | **75.6** | **65.1** | **80.2** | **67.3** |
|  | symmetric | 85.1 | 78.3 | 88.2 | 80.9 | 68.9 | 58.3 | 74.0 | 63.3 |

Table S7. **Kernel Symmetrization.**

merged, and finally pass through the remaining blocks. As in many studies [1, 9, 17] utilizing local information, using the output of $f_3$ as an input results in the best performance; thus, we select the feature map from $f_3$ as the input of the re-ranking network.

**Scale selection (Tab. S6).** We conduct experiments with a selection of scales to construct a multi-scale feature pyramid. Note that a high-scale feature can be helpful in terms of performance. However, considering the limitation of time and memory, we only scale the feature to a lower scale. The results show that constructing a cross-scale correlation using several scales has a clear performance advantage over the single-scale feature correlation method. Based on the experimental results, we finally select $S = 3$ scales.

**Symmetric kernel (Tab. S7).** Image similarity is essentially permutation-invariant, except in special cases. When we train a 4D convolutional network to predict image similarity, we can induce the network to be permutation-invariant in several ways. For instance, we can set the loss function to ensure that the output does not vary regardless of the input order. Another method is to make the 4D convolution kernel symmetrical. We experiment with the latter case as shown in Tab. S7. However, forcing the kernel to be symmetric did not yield good performance. Therefore, in this study, we softly induce permutation-invariant properties in the re-ranking network using loss symmetrization.

**Score fusion weight (Fig. S3).** To simultaneously verify the global and local relationships between two images, we re-rank the retrieval results based on the combined score $s_g + \alpha s_r$, where $s_g$ is the cosine similarity of the global de-

| # | layer | Medium | | | | Hard | | | |
|---|---|---|---|---|---|---|---|---|---|
| | | $\mathcal{R}$Oxf | +1M | $\mathcal{R}$Par | +1M | $\mathcal{R}$Oxf | +1M | $\mathcal{R}$Par | +1M |
| 0 | Global | 81.0 | 72.6 | 88.8 | 79.0 | 62.1 | 50.2 | 76.5 | 60.2 |
| 0 | $\alpha$QE | 85.4 | 77.5 | **90.7** | 83.5 | 67.5 | 57.8 | 79.8 | 66.9 |
| 100 | CV | 86.1 | 77.6 | 89.4 | 79.9 | 72.8 | 61.1 | 78.6 | 63.9 |
| 200 | CV | 87.2 | 78.9 | 90.0 | 81.2 | 74.5 | 62.9 | 79.5 | 66.0 |
| 400 | CV | **87.9** | **80.7** | 90.5 | 82.4 | **75.6** | **65.1** | **80.2** | **67.3** |
| 0 | $\alpha$QE | 85.4 | 77.5 | 90.7 | 83.5 | 67.5 | 57.8 | 79.8 | 66.9 |
| 100 | $\alpha$QE + CV | 88.0 | 80.5 | 90.9 | 84.1 | 74.6 | 65.2 | 80.9 | 70.0 |
| 200 | $\alpha$QE + CV | 88.8 | 82.1 | 91.2 | 84.9 | 75.9 | 67.4 | 81.6 | 71.5 |
| 400 | $\alpha$QE + CV | **89.3** | **82.8** | **91.6** | **85.3** | **77.1** | **68.6** | **82.2** | **70.7** |

Table S8. **Comparison with alphaQE.**

scriptors, $s_r$ is the output score of the re-ranking network, and $\alpha$ is the given weight for the re-ranking network output score $s_r$. Parameter $\alpha$ is tuned in $\mathcal{R}$Oxf/$\mathcal{R}$Par and fixed for a large-scale experiment and GLDv2-retrieval test, as in previous studies [1, 8, 12, 15]. Fig. S3a and Fig. S3b shows $\mathcal{R}$Oxf-Hard/$\mathcal{R}$Par-Hard performances according to score fusion weight $\alpha$. In these results, the re-rank score significantly improves the retrieval performance even if an extremely small re-rank score is added to the global descriptor matching score. Moreover, the best performance corresponded to an $\alpha$ value of 0.5. Based on these experimental results, we set $\alpha = 0.5$ for the re-ranking process.

### S2.4. Comparison with Query Expansion

**Comparison with $\alpha$QE (Tab. S8).** This study focuses on improving the image matching ability for single pairs. Therefore, we have not considered certain re-ranking methods such as diffusion [2, 5] or query expansion [3, 11], which require additional expenses to traverse the entire database mentioned in the main body of this paper. Although we do not consider them because of their different scopes, in this subsection we show that these re-ranking methods and the proposed re-ranking method can be harmoniously fused when they are used. Specifically, we compared and fused CV with one of the representative query expansion methods: $\alpha$-weighted query expansion ($\alpha$QE).

In contrast to geometric verification (GV) or our proposed correlation verification (CV), which evaluates the similarity between *two images*, the query expansion aggregates the query itself and its top-ranked neighbors *across the dataset* and creates an aggregated query to perform re-ranking. In the $\alpha$QE method, aggregation is performed with weighted averaging, and the weight of the $i$th ranked image is given by $(\mathbf{d}_q \cdot \mathbf{d}_i)^{\alpha_{QE}}$, where $\mathbf{d}_q$ is the global descriptor of the query image and $\mathbf{d}_i$ is the global descriptor of the $i$th ranked image for the query. Finally, the aggregated query descriptor $\mathbf{d}'_q$ is computed as follows:

$$\mathbf{d}'_q = \frac{\mathbf{d}_q + \sum_{i=1}^{n}((\mathbf{d}_q \cdot \mathbf{d}_i)^{\alpha_{QE}} \cdot \mathbf{d}_i)}{1 + \sum_{i=1}^{n}(\mathbf{d}_q \cdot \mathbf{d}_i)^{\alpha_{QE}}}, \quad (S1)$$

where $n$ is the number to aggregates, and $\alpha_{QE}$ is a pa-

| K | Medium | | | | Hard | | | |
|---|---|---|---|---|---|---|---|---|
| | $\mathcal{R}$Oxf | +1M | $\mathcal{R}$Par | +1M | $\mathcal{R}$Oxf | +1M | $\mathcal{R}$Par | +1M |
| 576 | 77.5 | 70.0 | **89.8** | 78.0 | 55.7 | 44.6 | 77.9 | 58.9 |
| 4608 | 78.5 | 71.3 | **89.8** | 78.7 | 58.1 | 46.4 | **78.3** | 59.3 |
| 73728 | **81.0** | **72.6** | 88.8 | **79.0** | **62.1** | **50.2** | 76.5 | **60.2** |

Table S9. **Queue Size of Momentum Contrastive Loss.**

| Loss | Medium | | | | Hard | | | |
|---|---|---|---|---|---|---|---|---|
| | $\mathcal{R}$Oxf | +1M | $\mathcal{R}$Par | +1M | $\mathcal{R}$Oxf | +1M | $\mathcal{R}$Par | +1M |
| SupCon [6] (Eq. (S2)) | 79.9 | 72.1 | **89.4** | 78.9 | 59.1 | 48.6 | **77.4** | 59.3 |
| Ours (Eq. (S3)) | **81.0** | **72.6** | 88.8 | **79.0** | **62.1** | **50.2** | 76.5 | **60.2** |

Table S10. **Comparison with SupCon Loss.**

rameter that amplifies or reduces the weight. $n$ and $\alpha_{QE}$ are tuned in $\mathcal{R}$Oxf/$\mathcal{R}$Par over the ranges: $n \in [1, 20]$ and $\alpha_{QE} \in [0.1, 2.0]$ and we finally set them to 5 and 2.0, respectively. Tab. S8 shows the re-ranking results using $\alpha$QE, the re-ranking results using CV, and the re-ranking results using $\alpha$QE and CV sequentially. For all settings, CV exhibits performance that is superior to the $\alpha$QE method, and even more superior when fused with the $\alpha$QE method.

### S2.5. Momentum Contrastive Loss Analysis

**Queue size (Tab. S9).** Our global backbone network, CVNet-Global, constructs a queue to store and leverage numerous samples for contrastive learning. Because queue size is one of the crucial factors that is directly related to the number of contrastive samples, we conduct experiments by varying the queue size $K$. Tab. S9 shows performances for different queue sizes. Overall, our global model benefits from a large $K$ value. A large queue size implies that several contrastive samples can be utilized, which can lead the global model to learn a more generalized representation.

**Differences in SupCon loss (Tab. S10).** Our momentum contrastive loss is similar to SupCon [6] loss. Similar to the SupCon loss, it performs contrast learning with multiple positives using labels. The primary difference between these losses is that the SupCon loss assumes a relatively constant number of positives. However, a large difference exists in the number of positives for each sample because of the class imbalance data and queue structure. In SupCon loss (Eq. (S2)), because all query-positive cosine similarities are included in the denominator, the scale of the loss is significantly affected by the number of positives:

$$\mathcal{L}_s = \frac{-1}{|P(q)|} \sum_{p \in P(q)} \log \frac{\exp\left(\bar{\mathcal{C}}\left(\mathbf{d}_g^q \cdot \bar{\mathbf{d}}_g^p, 1\right)/\tau\right)}{\sum_{i \in P(q) \bigcup N(q)} \exp\left(\bar{\mathcal{C}}\left(\mathbf{d}_g^q \cdot \bar{\mathbf{d}}_g^i, \mathbb{1}_q^i\right)/\tau\right)}.$$
(S2)

To solve this scale problem, we design our contrastive loss (Eq. (S3)) similar to the SupCon loss $\mathcal{L}_s$. However, only the target positive $p$ is included in the denominator.

| model | Medium | | | | Hard | | | |
|---|---|---|---|---|---|---|---|---|
| (Rerank top-100) | $\mathcal{R}$Oxf | +1M | $\mathcal{R}$Par | +1M | $\mathcal{R}$Oxf | +1M | $\mathcal{R}$Par | +1M |
| R50-DELG[†] | 71.1 | 60.4 | 86.9 | 70.9 | 47.0 | 32.0 | 73.6 | 48.1 |
| + CVNet-Rerank | **78.7** | **67.7** | **87.9** | **72.3** | **63.0** | **46.1** | **76.8** | **52.5** |
| R50-DOLG[†] | 79.0 | 70.0 | 88.3 | 76.2 | 57.5 | 43.2 | 75.0 | 55.4 |
| + CVNet-Rerank | **83.7** | **74.9** | **89.0** | **77.2** | **69.1** | **55.7** | **77.1** | **59.0** |

Table S11. **Performance with Different Backbones.**

| model (R50) | Medium | | | | Hard | | | |
|---|---|---|---|---|---|---|---|---|
| (Rerank top-100) | $\mathcal{R}$Oxf | +1M | $\mathcal{R}$Par | +1M | $\mathcal{R}$Oxf | +1M | $\mathcal{R}$Par | +1M |
| CVNet-Global | 81.0 | 72.6 | 88.8 | 79.0 | 62.1 | 50.2 | 76.5 | 60.2 |
| + CorrNet | 81.3 | 72.7 | 88.8 | 79.0 | 62.3 | 50.3 | 76.5 | 60.2 |
| + HNM | 84.6 | 75.9 | 89.0 | 79.3 | 69.3 | 57.0 | 76.9 | 61.1 |
| + CSC | 85.8 | 77.5 | 89.3 | **79.9** | 71.6 | 60.5 | 78.1 | 63.7 |
| + HaS | **86.1** | **77.6** | **89.4** | **79.9** | **72.8** | **61.1** | **78.6** | **63.9** |

Table S12. **Module Ablation Study.**

$$\mathcal{L}_{con} = \frac{-1}{|P(q)|} \sum_{p \in P(q)} \log \frac{\exp\left(\bar{\mathcal{C}}\left(\mathbf{d}_g^q \cdot \bar{\mathbf{d}}_g^p, 1\right)/\tau\right)}{\sum_{i \in \{p\} \bigcup N(q)} \exp\left(\bar{\mathcal{C}}\left(\mathbf{d}_g^q \cdot \bar{\mathbf{d}}_g^i, \mathbb{1}_q^i\right)/\tau\right)}.$$
(S3)

Tab. S10 shows the results of training CVNet-Global using each of the two losses. When using the proposed contrastive loss $\mathcal{L}_{con}$, the results are more stable than when using SupCon loss $L_s$.

### S2.6. Performance with Different Backbones

Tab. S11 shows the results when CVNet-Rerank is combined with the standard global models, DELG and DOLG (both reproduced[†]). Both models are trained using the settings of the original paper except for setting the maximum number of epochs to 25 epochs. Afterward, the proposed CVNet-Rerank is trained with each global backbone. As shown in Tab. S11, CVNet-Rerank also works well when combined with other global backbones.

### S2.7. Module Ablation Study

Tab. S12 shows the results when the components of CVNet-Rerank are added to CVNet-Global one by one. From the Tab. S12, we can see that when the Correlation encoding Network (CorrNet) is added, the accuracy is slightly improved but when Hard Negative Mining (HNM) is applied, the accuracy is significantly improved. From the observation, we can see that the correlation encoding network and hard negative mining is a good combination but the network is quite hard to train if the hard negative mining is not used. Based on our experience, when we train the network on random negatives without hard negative mining, the training is dominated by the very rare high correlations between the query and the random negatives degrading the accuracy. To prevent the degradation, we have to train the network using hard negative mining. In summary, we can say that the correlation encoding network and hard negative mining is the core component of the CVNet-Rerank, and

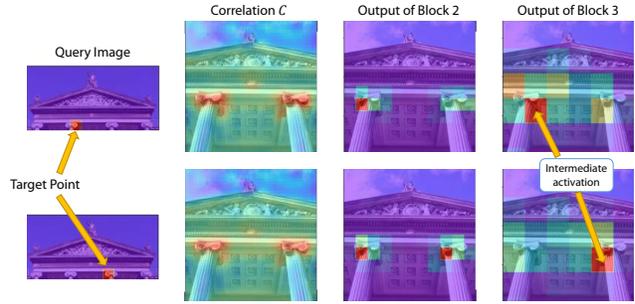

Figure S4. **Intermediate Feature Visualization.** Our network naturally learns the correct geometric relationship of dense matching and pays attention to the correct position by compressing the surrounding matching information from the 4D correlation.

the combination significantly improves the accuracy. Cross-Scale Correlation (CSC) and Hide-and-Seek (HaS) are the optional choices that can incrementally improve the accuracy of the re-ranking network.

## S3. Intermediate Feature Visualization

We visualize the intermediate features of our re-ranking model to see how the network interprets and compresses the correlation. To visualize the intermediate 4D features, we select one target point from the query side and visualize the magnitude of the corresponding feature parts on the key side. The visualized results are presented in Fig. S4. In Fig. S4, we observe that the model focus on the correct position by compressing the surrounding matching information from the 4D correlation. As shown in the results, the network naturally learns the geometric pattern of dense matching without any predefined geometric model (*e.g.* Affine model). Additional intermediate feature visualizations are presented in Fig. S5.

## S4. Reproducing Details

For a fair comparison with other re-ranking methods, we conduct experiments by reproducing other re-ranking methods based on the global backbone network. We reproduce two re-ranking methods: geometric verification (GV) and Reranking Transformer [14]. Because both methods are based on the local features of DELG, we attach the local branch of DELG [1] to our global backbone (R50-CVNet-Global) to learn the local features of DELG. All local-feature-related settings are identical to those in the DELG [1]. During testing, we extract a maximum of 1000 local features (500 for RRT) and use them for the re-ranking process.

**Geometric Verification (GV).** We reproduce the GV based on the DELG. Official code of DELG uses RANSAC [4], which belongs to the scikit-learn [10] package; however, we could not improve the re-ranking performance with

this version. Finally, we implement RANSAC using pydegensac [7], which exhibits performance superior to that of scikit-learn. Additionally, as mentioned in DELG [1] paper, we set a minimum number of inliers to improve re-ranking performance. We tune the minimum number of inliers over the range: [10,300], and finally set it to be 100.

**Reranking Transformers (RRT).** We train the RRT model with official code provided by [14], and use all the same settings are as the provided one. The only difference is that we input the global descriptor extracted from CVNet-Global and local features extracted from the added local branch, instead of the features extracted by the pre-trained DELG model.

## S5. Additional Qualitative Results

Additional qualitative results on $\mathcal{R}$Oxford5k-Hard+1M, $\mathcal{R}$Paris6k-Hard+1M, and the GLDv2-retrieval-test are shown in Fig. S6, Fig. S7, and Fig. S8, respectively. The results show that the proposed re-ranking method performs re-ranking robustly, even if the global descriptor matching results in misjudgment in situations involving challenges such as viewpoint change, occlusion, and truncation.

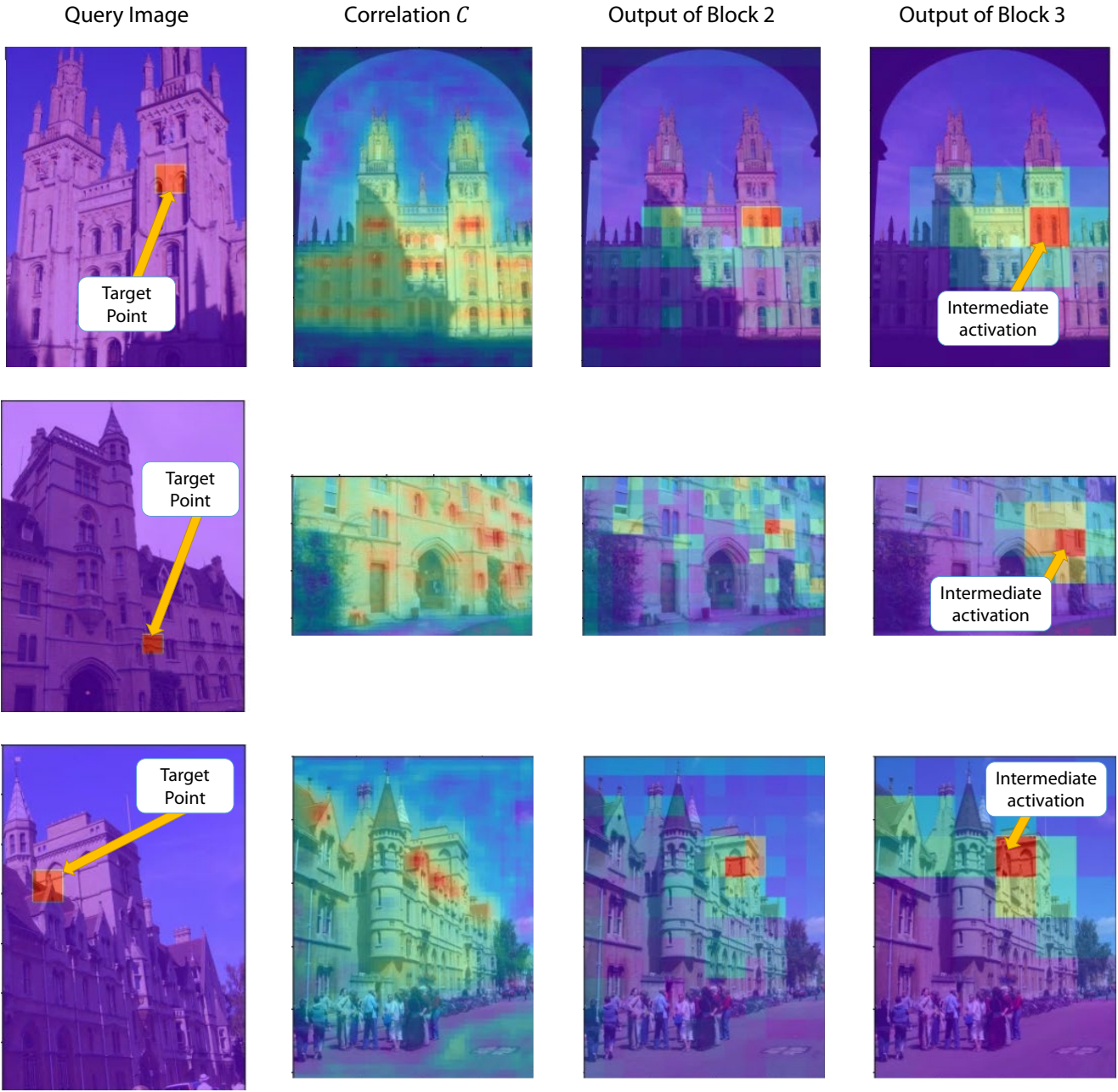

Figure S5. **Additional Intermediate Feature Visualization.**

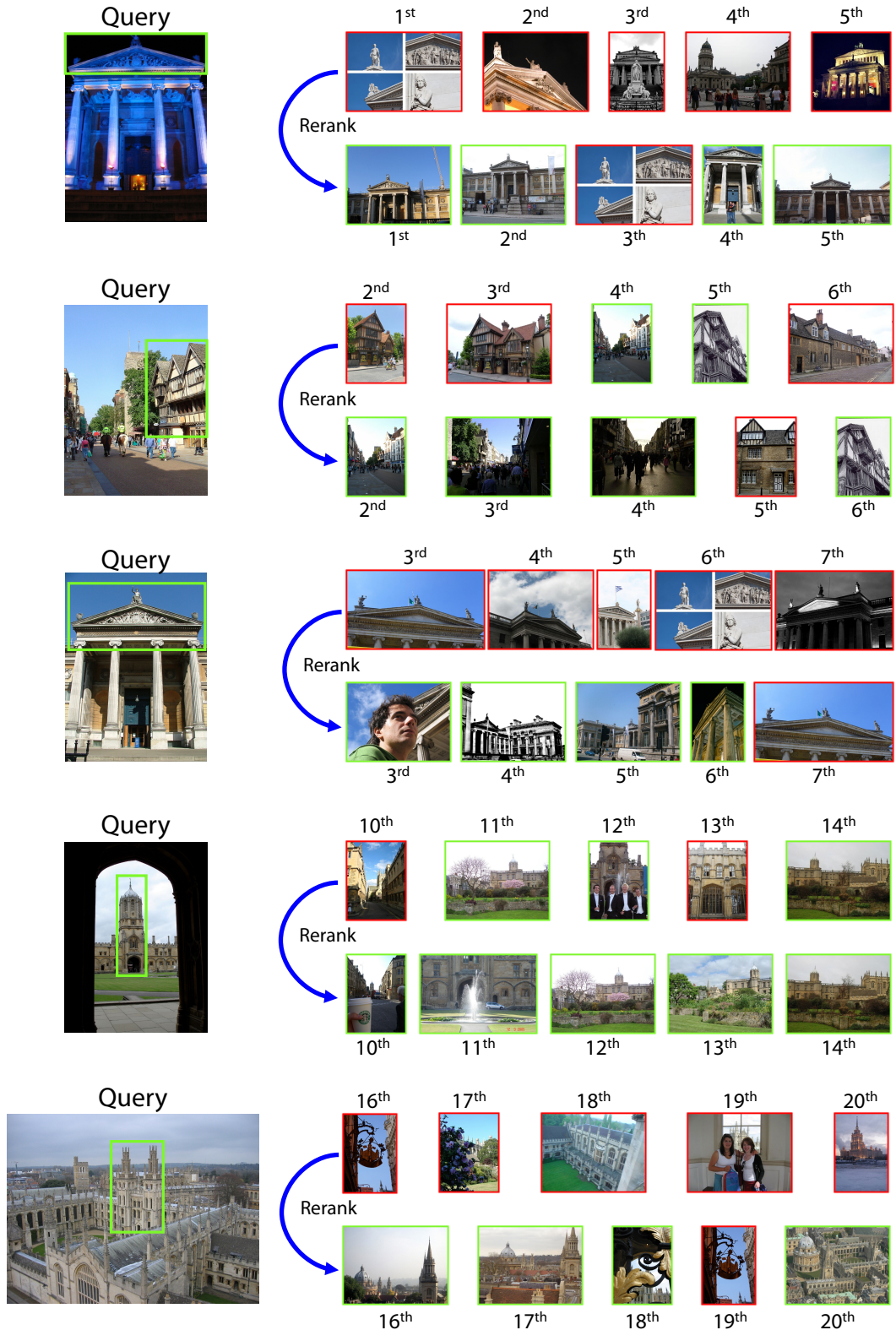

Figure S6. **Additional Qualitative Results on $\mathcal{R}$Oxford5k-Hard+1M with R50-CVNet.** The upper line is the global descriptor matching result and the lower line is the re-ranking result. Correct/incorrect results are marked with green/red borders, respectively. The query used as an input is generated by cropping only the part bounded by a green square. Our purpose is to visualize the difference between global descriptor matching and re-ranking, so we skip the results of the ranks that are correct in both the global descriptor matching and re-ranking processes.

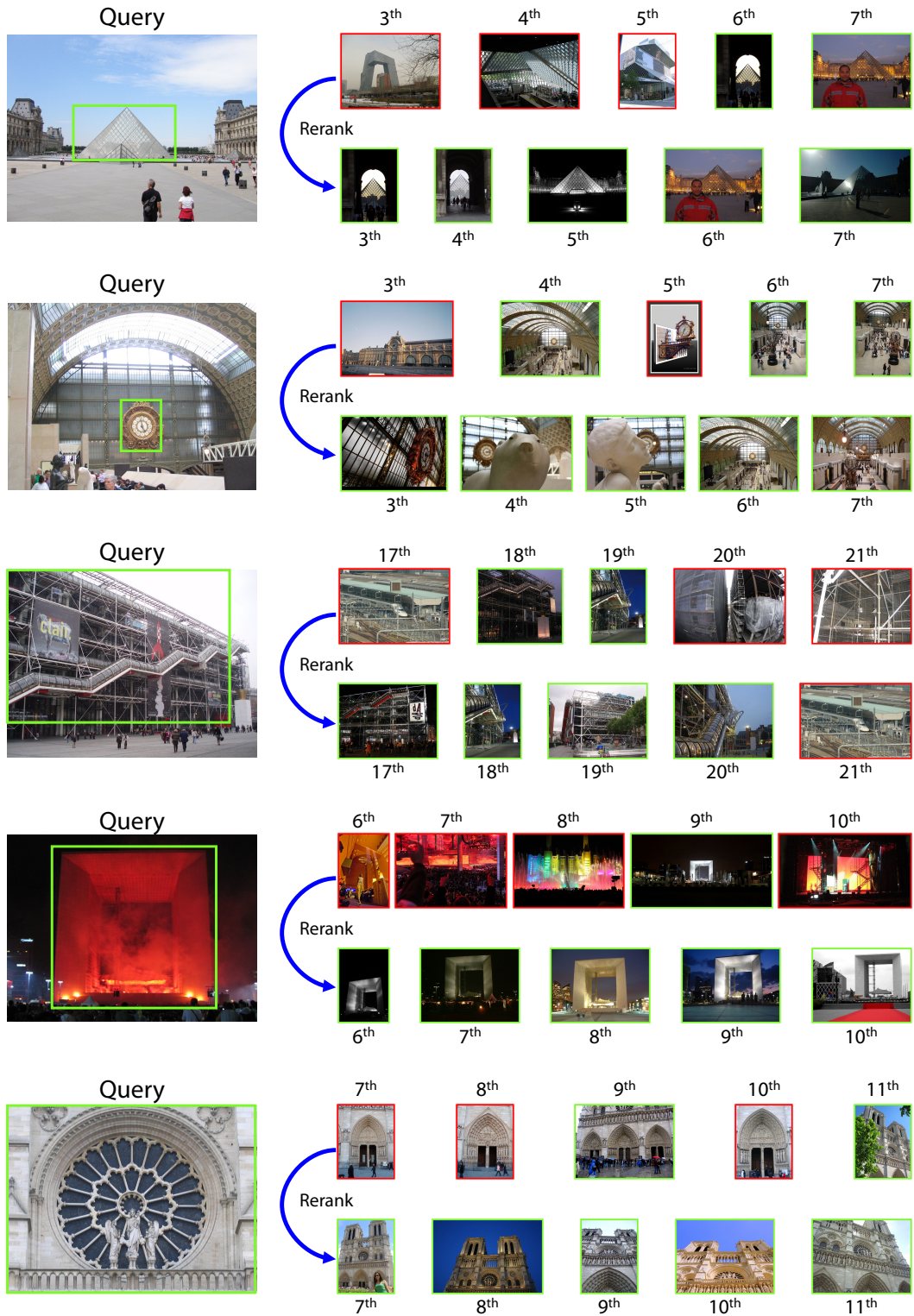

Figure S7. **Additional Qualitative Results on $\mathcal{R}$Paris6k-Hard+1M with R50-CVNet.** The upper line is the global descriptor matching result and the lower line is the re-ranking result. Correct/incorrect results are marked with green/red borders, respectively. The query used as an input is generated by cropping only the part bounded by a green square. Our purpose is to visualize the difference between global descriptor matching and re-ranking, so we skip the results of the ranks that are correct in both the global descriptor matching and re-ranking processes.

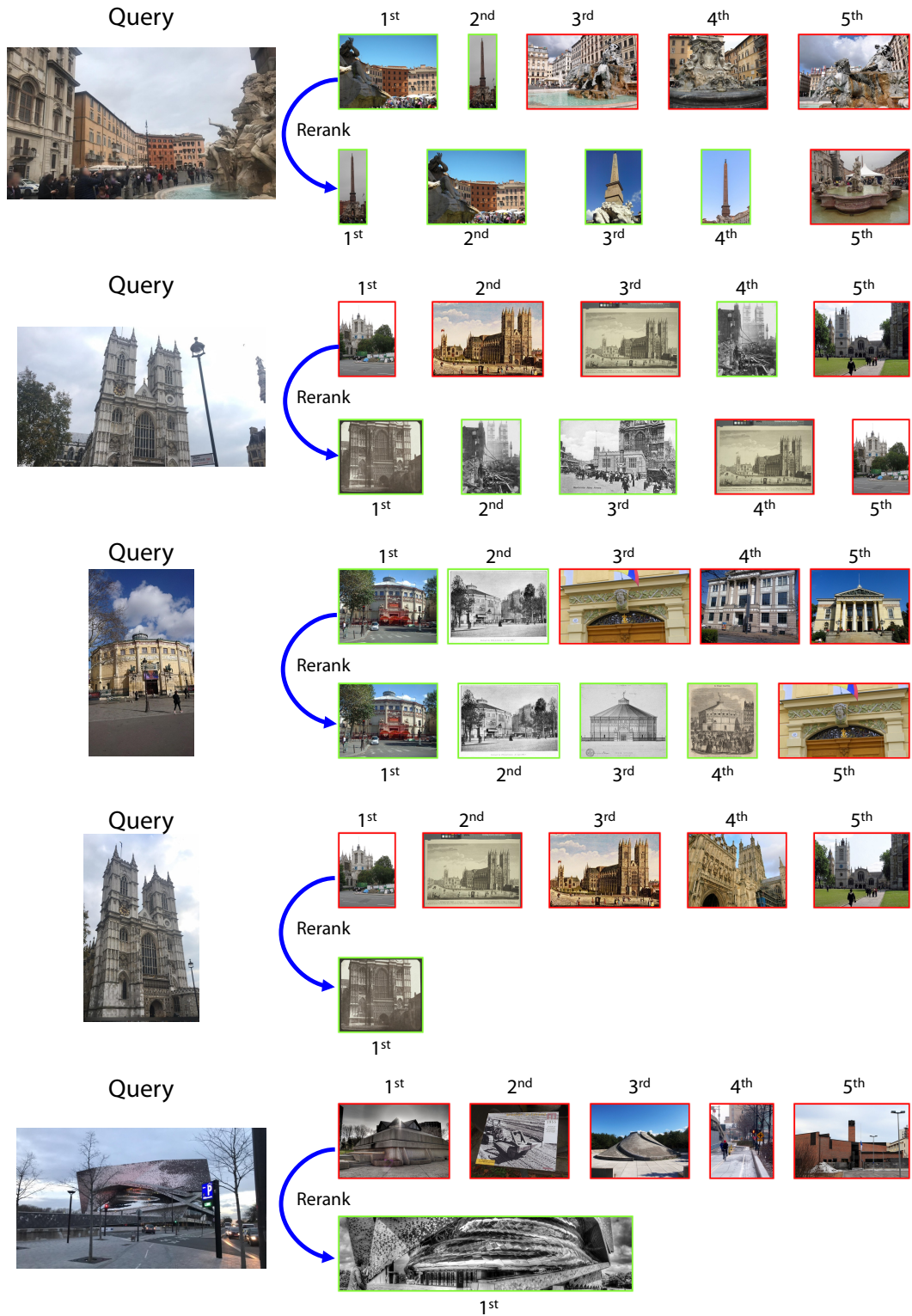

Figure S8. **Qualitative Results on GLDv2-retrieval-test with R50-CVNet.** The upper line is the global descriptor matching result and the lower line is the re-ranking result. Correct/incorrect results are marked with green/red borders, respectively. The last two queries each have only one positive sample, so we skip the results after the correct answer comes out.


\end{document}